\documentclass[final]{cvpr}

\usepackage{times}
\usepackage{epsfig}
\usepackage{graphicx}
\usepackage{amsmath}
\usepackage{amssymb}
\usepackage{booktabs} 
\usepackage{graphbox}
\usepackage{multirow}

\usepackage{enumitem}
\usepackage{xr}

\usepackage[pagebackref=true,breaklinks=true,colorlinks,bookmarks=false]{hyperref}

\def\cvprPaperID{4327} 
\def\confYear{CVPR 2021}

\begin{document}

\title{HDR Environment Map Estimation for Real-Time Augmented Reality}

\author{Gowri Somanath\\
	Apple\\
	{\tt\small gowri@apple.com}
	\and
	Daniel Kurz\\
	Apple\\
	{\tt\small daniel\textunderscore kurz@apple.com}
}

\maketitle

\begin{abstract}
We present a method to estimate an HDR environment map from a narrow field-of-view LDR camera image in real-time. This enables perceptually appealing reflections and shading on virtual objects of any material finish, from mirror to diffuse, rendered into a real  environment using augmented reality. Our method is based on our efficient convolutional neural network, EnvMapNet, trained end-to-end with two novel losses, ProjectionLoss for the generated image, and ClusterLoss for adversarial training. Through qualitative and quantitative comparison to state-of-the-art methods, we demonstrate that our algorithm reduces the directional error of estimated light sources by more than $50\%$, and achieves $3.7$ times lower Frechet Inception Distance (FID). We further showcase a mobile application that is able to run our neural network model in under $9$ ms on an iPhone~XS, and render in real-time, visually coherent virtual objects in previously unseen real-world environments.

\end{abstract}


\begin{figure*}
	\centering
	\includegraphics[width=0.81\textwidth]{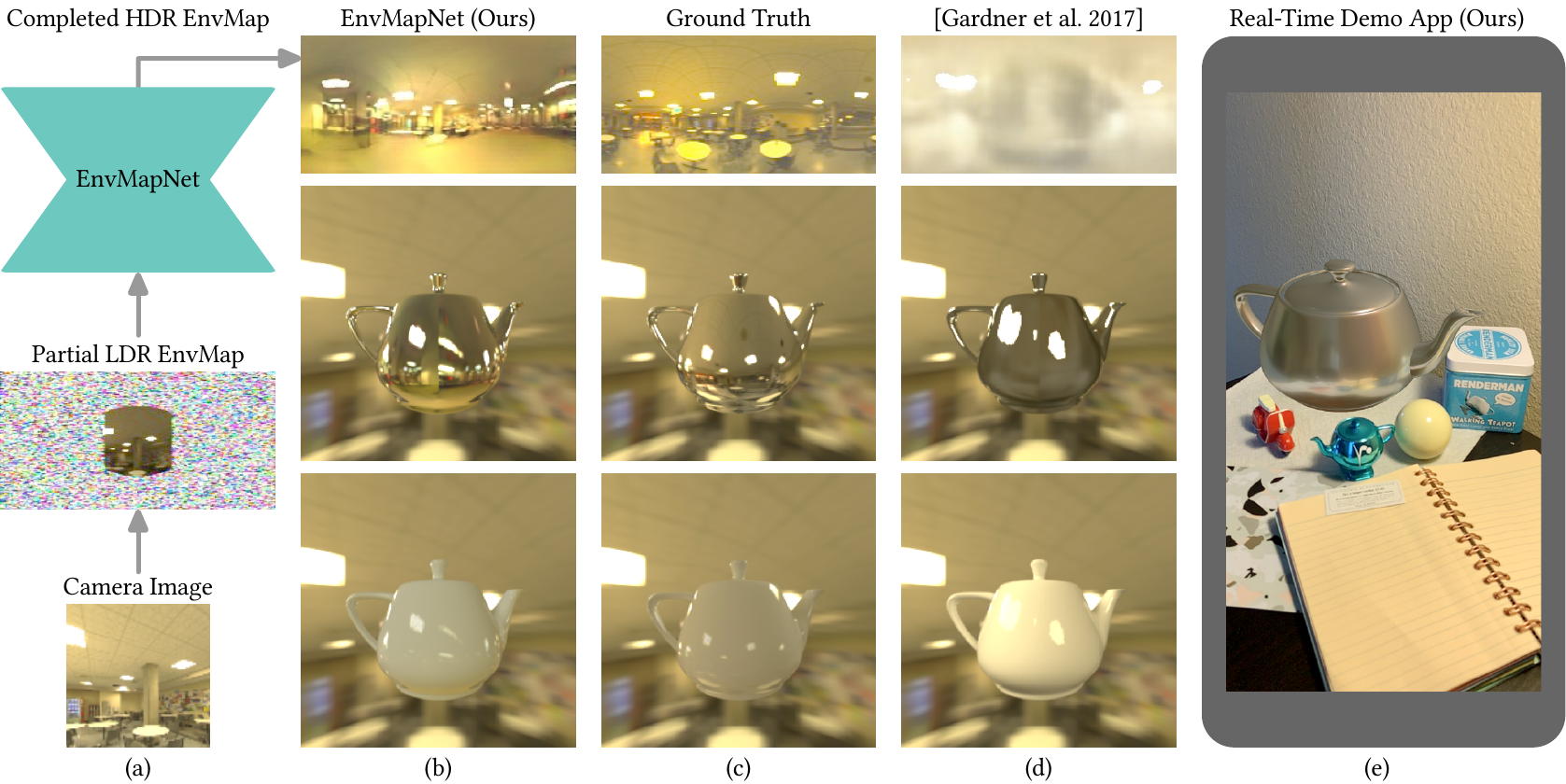} 
	\caption{Given a partial LDR environment map from a camera image (a), we estimate a visually coherent and completed HDR  map that can be used by graphics engines to create light probes. These enable rendering virtual objects with any material finish into the real environment in a perceptually pleasing and realistic way (b), more similar to ground truth (c) than state-of-the-art methods (d). Our mobile app renders a virtual  teapot using our estimated environment map in real-time (e). \textbf{See supplementary material for videos.} }
	\label{teaser-fig}
\end{figure*}

\setlength{\abovedisplayskip}{3pt}
\setlength{\belowdisplayskip}{3pt}
\section{Introduction}
In this work, we discuss video see-through augmented reality (AR) applications, in which virtual objects are superimposed on camera frames of the real environment shown on an opaque display, e.g. on a phone as shown in Fig.~\ref{teaser-fig}(e). Creating immersive and believable AR experiences involves many aspects of computer vision and  graphics. One of the requirements is visual coherence: the problem of matching visual appearance of rendered  objects to their real-world background, such that virtual and real objects become indistinguishable in the composited video. Accomplishing this involves matching various scene and camera properties, such as lighting, geometry, and sensor noise.

This paper focusses on creating reflections and lighting  for virtual objects by estimating an omnidirectional HDR environment map. To support rendering objects with a variety of geometry, material properties, and dimensions, the environment map must be high dynamic range, and have sufficient image resolution to represent objects and features in the scene. We use the equirectangular projection and RGB color space for the environment maps. As shown in Fig.~\ref{teaser-fig}(a), the  challenge in mobile AR is limited camera field of view (FoV) and motion by the user, hence an application is usually able to accumulate less than $100$ degrees effective FoV. A virtual object placed in front of the user, however, is expected to reflect what is behind the camera, and parts which are not present in the captured frames. The problem is thus to estimate, given this incomplete environment map, a plausible estimation for  rest of the scene and its lighting.
We show that our method is not only able to estimate the light information, but also to synthesize a high resolution completed scene. For instance in the scene shown in Fig.~\ref{teaser-fig}(b), the estimated environment map is high resolution, continuous, and a plausible extrapolation of the input. The synthesized parts not only match low frequency information (ambient light temperature and  intensity), but also finer details such as the type of light sources (in this case, ceiling area lights). As detailed in Sec.~\ref{sec_method}, we achieve this context-aware scene completion using the framework of generative adversarial networks (GANs)~\cite{GoodfellowGAN2014} along with novel loss functions, ProjectionLoss and ClusterLoss, designed for accurate light estimation. 

In mobile AR frameworks \cite{arcore,arkit}, we can obtain camera frames, poses and scene geometry, around the 3D location where the virtual object is to be placed. This allows a real-time renderer to create light probes \cite{refprobesunity} at the 3D location. RGB texture information from the  frames can  be rendered into an equirectangular image at these probe locations. In this work, we focus on processing the partial environment map at these probes. Our method takes as input a partial environment map that is composed from one or more low dynamic range (LDR) camera frames (8 bit per channel), and outputs a completed environment map that is higher dynamic range (HDR, 16 bit channel). Thus we perform both lifting of input pixels from LDR to HDR, as well as spatial HDR image extrapolation. The output environment map retains the color and details from pixels that were in the input, while filling the unknown pixels with  plausible content that is coherent with the known. 
That is, we want the completed environment map to represent the textures from a plausible real scene. Through detailed quantitative and qualitative comparisons, we demonstrate that our method surpasses the current state-of-the art to estimate high quality, perceptually plausible, and accurate HDR environment maps. We reduce the directional (angular) error for lights by more than half, and achieve a significantly lower Frechet Inception Distance (FID). 

Related works estimate a subset of the properties, such as lighting direction, color, and intensity using low resolution images~\cite{Garon_2019_CVPR,Gardner_2019_ICCV,LeGendre:2019:DLI:3306307.3328173},   parametric lights~\cite{holdgeoffroy-cvpr-17,Gardner_2019_ICCV}, or LDR environment maps~\cite{srinivasan20lighthouse}. Methods that estimate both light and texture \cite{gardner-sigasia-17,LeGendre:2019:DLI:3306307.3328173,song2019neural} lack resolution and detail in the estimated environment map. In contrast, our solution provides sufficient details in the image for virtual objects of any material finish including reflective mirrors. To create plausible scene completions, we take inspiration from work in the area GANs for image synthesis. However, most success for GANs has come for datasets with single objects (e.g. faces~\cite{karras2018progressive}), using conditional labels~\cite{pmlr-v70-odena17a}, or self-supervision~\cite{Chen2018SelfSupervisedGV}. Its direct use for panoramic images of indoor scenes, small datasets with high appearance variation, and ambiguity in semantic labeling, is challenging. We present ClusterLoss, a novel training loss for the discriminator, that allows us to create realistic images with a dataset of only $\sim$2,800 training images.

We also observe that each of the previous methods have a different evaluation scheme, are based on respective representations and/or involve subjective user studies. Though subjective plausibility is important for user experience, it makes benchmarking difficult. 
We present metrics that can be used to quantitatively compare both the lighting and reflection quality of environment maps.

 In summary, we make the following core contributions:
\begin{itemize}[noitemsep,nolistsep]
	\item We present a method to generate an HDR environment map suitable for both reflections and lighting from a small FoV LDR image or partial environment map. To our knowledge, ours is the first to achieve this for real-time AR applications (under $9$ms on iPhone~XS).
	\item We present two novel contributions to the training pipeline in the form of ProjectionLoss for the environment map, and ClusterLoss in the adversarial training. This allows us to reduce the directional error by more than $50\%$ compared to current methods, while achieving $3.7$ times lower FID.
	\item We establish metrics that allow easy quantitative comparisons with related work, and thus provide a systematic benchmark for this emerging area in AR.

\end{itemize}

\section{Related work}
\label{sec:related}
A classical technique to obtain an HDR environment map is to merge images of a mirror sphere in the scene captured under multiple exposure brackets~\cite{debevec98}. This can be applied for offline use cases but is unsuitable for real-time mobile AR  in arbitrary and novel environments. Some previous works have used additional cues about an object~\cite{Marschner1997InverseLF,portraitrelighting}, scene geometry~\cite{azinovic2019inverse,maier2017intrinsic3d,zhang2016emptying}, or special cases such as  sun position estimation~\cite{outdoorcvpr19,holdgeoffroy-cvpr-17,deepsky2019}.  For brevity, we only discuss works that focus on light estimation from small FoV images of general scenes. In this context, works can be divided into two categories: those that focus on light estimation with low dimensional parameters, and those which estimate both lighting and environment maps.

Garon~\etal~\cite{Garon_2019_CVPR} use spherical harmonics representation for light and depth estimated using the SUNCG dataset \cite{song2016ssc}\footnote{SUNCG is currently withdrawn from distribution}. Cheng~\etal~\cite{dachuan2018} also predict $48$  spherical harmonics coefficients from two images captured by the front and rear cameras of a mobile device. Gardner~\etal use parametric lights as representation \cite{Gardner_2019_ICCV}. The parameters are derived using peak finding, region growing, and ellipse fitting on intensity images. Even though use of parametric lights reduces the decoder size, the authors use L2 loss by converting the parameters to an equirectangular image. It is not clear how to extend this method to generate higher quality texture in the environment map that would be consistent with the regressed parameters. We thus choose to have an end-to-end network that directly estimates the HDR environment map. We use the same parametric lights to create quantitative metrics for benchmarking.

Gardner~\etal use equirectangular representation \cite{gardner-sigasia-17} by dividing the task into light position estimation (trained with LDR panoramas), and HDR intensity estimation (trained with Laval HDR Images dataset). This is the work most similar to our method in input-output, representation, and formulation. In contrast to their method, we have a single stage training from an LDR small FoV image to a completed HDR environment map, and using our proposed adversarial training we are able to generate more realistic RGB scene completion for use on reflective virtual objects. Some recent works, like ours, employ an adversarial loss to generate a completed environment map  \cite{song2019neural, srinivasan20lighthouse} or sphere images \cite{LeGendre:2019:DLI:3306307.3328173}. LeGendre~\etal captured a special dataset with three spheres of mirror, matte, and diffuse gray finish \cite{LeGendre:2019:DLI:3306307.3328173}. They train a network  to regress from a camera image to three small ($32\times32$ pixel) images representing the sphere segments. This  precludes use of the results for high quality reflections. Also, due to the low resolution, the estimate covers mostly the portion of the scene behind the camera. It is useful to render a given frame, but as the camera commonly moves with respect to the virtual object in a AR application, a new estimate has to be inferred frequently. Thus even with the assumption of a static scene, their method would suffer from temporal flickering due to multiple independent estimates.

Song~\etal use a multi-stage ensemble that estimates geometry, LDR completion and  HDR illumination \cite{song2019neural}. Our method uses a single model, and we do not require depth map per HDR image for training. This makes our method efficient to train and use in real-time mobile AR.  The authors do not specify the resolution of the panoramic images, but visual observation shows lack of resolution and artifacts due to the projection from noisy 3D reconstructions.
Srinivasan~\etal  use an input stereo pair to generate the output environment map~\cite{srinivasan20lighthouse}. They use the LDR images from the synthetically generated InteriorNet dataset~\cite{InteriorNet18} to train the light estimation, by application of inverse gamma on the tone mapped images. This limits the ability to learn realistic high dynamic range and accurate lighting, as also indicated by their results on mostly specular objects. In contrast, our method does not require stereo input, and estimates an HDR environment map that can be applied for lighting objects with wide range of materials from diffuse to mirror.

 Furthermore, we employ only 2,100 HDR images of the publicly available Laval HDR Images dataset of real environments, a much smaller dataset than the $4.06$ million non-public specialized sphere images used in~\cite{LeGendre:2019:DLI:3306307.3328173}, the withdrawn SUNCG used in~\cite{song2016ssc}, or the 200,000+ Matterport images in~\cite{song2019neural}. The lack of shared code, model weights and use of private datasets make it difficult to make comparisons to these recent methods, that also compare  to ~\cite{gardner-sigasia-17} as we do in our quantitative evaluations. However, we present qualitative comparisons to these works in Sec.~\ref{sec_results}.  

With the exception of \cite{LeGendre:2019:DLI:3306307.3328173}, that outputs a $32\times32$ low resolution sphere image, no other  method has been demonstrated  on  mobile devices.  To the best of our knowledge, we are the first to demonstrate real-time on-device generation of environment maps of high resolution and  image quality.

\begin{figure*}[!ht]
	\centering
	\includegraphics[width=0.85\textwidth]{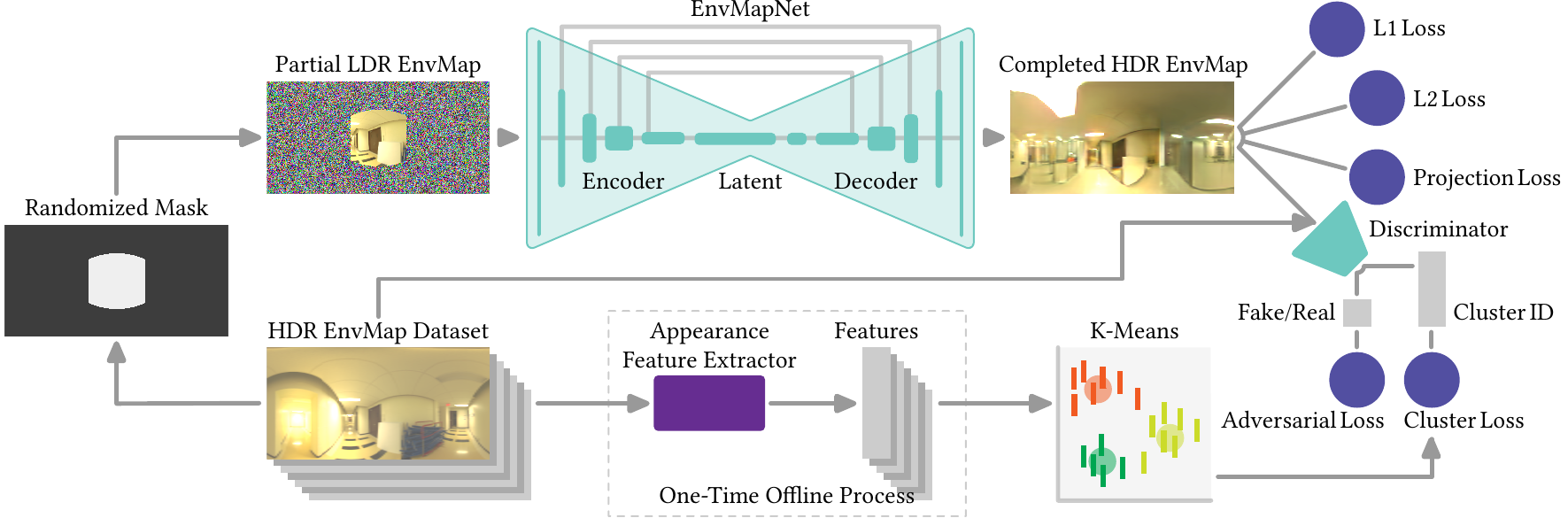}
	\caption{ Overview of our method: We propose EnvMapNet that estimates completed HDR environment maps from partial input. We train the network end-to-end in an adversarial setup. An offline one-time clustering of the training images is used to provide a supervised classification task for our novel ClusterLoss in the discriminator. The additional adversarial loss, along with our proposed ProjectionLoss, allows our method to generate high quality environment maps for reflection and accurate shading of virtual objects. }
	\label{modelarch}
\end{figure*}

\section{Proposed method}\label{sec_method}
Creating a light probe in a mobile AR application involves two broad stages: first to select a 3D point as center of this probe and project known scene information to an equirectangular environment map with the selected point as camera center; and second to process the partial equirectangular to output a completed HDR environment map. The first requires color and scene geometry knowledge or assumption for projection. In our work, we assume without loss of generality that platform and application-dependent processing can be used for this projection, and obtain the incomplete environment map from the probe center. Using the mobile device pose also helps to ensure that the environment maps are upright or gravity aligned. That is, the floor always appears at the bottom and the ceiling on top. Our focus in this paper is on completion, and LDR to HDR lifting of this incomplete environment map as shown in Fig.~\ref{modelarch}.

We use the equirectangular representation (128$\times$256 pixels)  and create a four channel input (RGB-mask). Known pixel intensities are normalized to $\mathnormal{[-1.0,1.0]}$, while the unknown pixels are populated with random noise from a uniform distribution $\mathnormal{U(-1.0,1.0)}$. The fourth channel is a binary mask with known pixels set to~$\mathnormal{0}$. This is input into our network, EnvMapNet, that outputs an HDR RGB image in log scale. The log image is converted to linear values, optionally decomposed into analytical lights, and provided to the renderer. We train our network end-to-end with  image-based  and adversarial losses,  which allows us to handle both the generative aspect for reflection completion as well as light estimation from partial environment map. 

\subsection{Dataset and  processing}\label{sec_data}
In this work we use two datasets: Laval Indoor HDR dataset~\cite{gardner-sigasia-17} and PanoContext LDR panoramas~\cite{panocontext}. We use the author's test split \cite{gardner-sigasia-17}\footnote{http://vision.gel.ulaval.ca/~jflalonde/projects/deepIndoorLight/test.txt} for HDR images, resulting in a total of 2,810 training images. The different sources and scenes have large intensity variations and hence are unsuitable for training as-is. We thus employ a log scale on exposure-normalized HDR linear images for the network. The LDR PanoContext images are only used to train the discriminator. For a given linear HDR ground truth image $\mathnormal{G_{lin}}$, we compute the training ground truth $\mathnormal{G}$ as:
 \begin{equation}
G = \min(\max(0, \log_{10}(G_{lin} \cdot \alpha + 1)),2) - 1, \\
 \end{equation}

We empirically choose $\alpha = 0.2 \cdot \overline{G_{lin}}$ to match the middle gray values. The clipping of intensities corresponds to our use of tanh activation for our network, and results in  dynamic range of $[0,100]$ for output linear RGB, given the input LDR images in $[0,1]$ range.

\subsection{Model architecture}\label{sec_model}
Our model, EnvMapNet, consists of an encoder and decoder with skip connections as shown in Fig.~\ref{modelarch}.  Each is composed of  building blocks detailed in Appendix~\ref{sec_network_details}. The encoder is composed of five sets of EnvMapNet-conv-block and EnvMapNet-downsample-blocks. The resulting latent vector is convolved with a 1$\times$1 kernel to output $64$ filters. The decoder mirrors the encoder by using EnvMapNet-conv-block and EnvMapNet-upsample-blocks.  The final output is produced by a  $3\times3$ convolution to produce $3$ channels for RGB, followed by a tanh activation. 

\begin{figure*}[!t]
	\centering
	\begin{tabular}{|c@{ }c||@{}c@{}|@{}c@{}|@{}c@{}|@{}c@{}|@{}c@{}|@{}c@{}|}
		\hline
		(a) & ProjectionLoss & 0.37 & 0.25 & 0.35 & 3.13 & 0.63 & 0.43\\
		\hline
		(b) & AngularError & 21.4 & 22.9 & 24.2  & 103.9 & 107.2 & 109.86 \\
		\hline
		(c) &\includegraphics[align=c,width=0.11\linewidth]{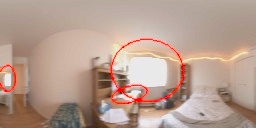} &
		\includegraphics[align=c,width=0.11\linewidth]{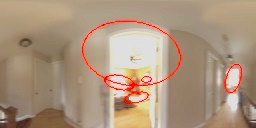} &
		\includegraphics[align=c,width=0.11\linewidth]{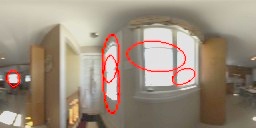} &
		\includegraphics[align=c,width=0.11\linewidth]{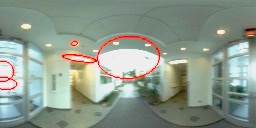} &
		 	\includegraphics[align=c,width=0.11\linewidth]{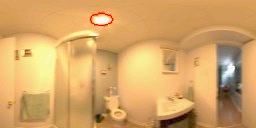} &
		\includegraphics[align=c,width=0.11\linewidth]{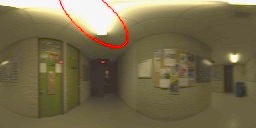} &
	 \includegraphics[align=c,width=0.11\linewidth]{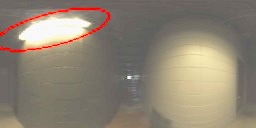}\\

		\hline
		(d) &\includegraphics[align=c,width=0.11\linewidth,trim={ 2.1cm 2.1cm 2.1cm 2.1cm},clip]{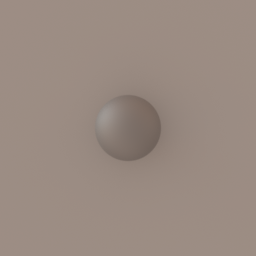} &
		\includegraphics[align=c,width=0.11\linewidth,trim={ 2.1cm 2.1cm 2.1cm 2.1cm},clip]{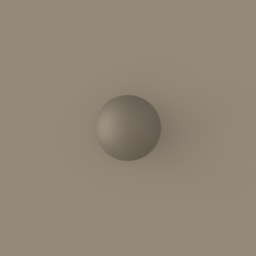} &
		\includegraphics[align=c,width=0.11\linewidth,trim={ 2.1cm 2.1cm 2.1cm 2.1cm},clip]{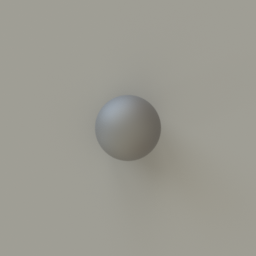} &
		\includegraphics[align=c,width=0.11\linewidth,trim={ 2.1cm 2.1cm 2.1cm 2.1cm},clip]{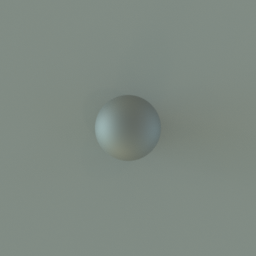} &
			\includegraphics[align=c,width=0.11\linewidth,trim={ 2.1cm 2.1cm 2.1cm 2.1cm},clip]{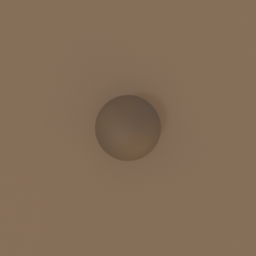}  &
		\includegraphics[align=c,width=0.11\linewidth,trim={ 2.1cm 2.1cm 2.1cm 2.1cm},clip]{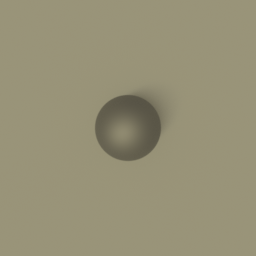} &
	\includegraphics[align=c,width=0.11\linewidth,trim={ 2.1cm 2.1cm 2.1cm 2.1cm},clip]{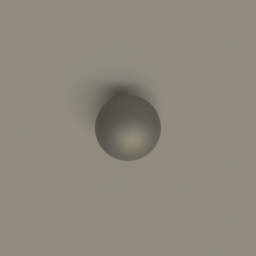}\\
		&	Reference  & Example 1 & Example 2 & Example 3 & Example 4 & Example 5 & Example 6\\
		\hline

	\end{tabular}
	\caption{Representative results for metrics and losses correlated to directional error of estimated lights.  Row (a):  Proposed projection loss on the environment maps w.r.t Reference (see Sec.~\ref{sec_regloss}). Row (b): Angular errors in degrees, using parametric lights for each environment map w.r.t Reference (see Sec.~\ref{sec_metrics}). Row (c): Environment maps with extracted parametric lights shown as red ellipses. Row (d):  Rendering of a rough metallic sphere using the corresponding environment map and viewed from top.   }
	\label{angularerrorsamples}
\end{figure*}

\subsection{Image-based  losses  and ProjectionLoss} \label{sec_regloss}
We use a weighted combination of image-based  and adversarial losses to train our model end-to-end. We want to retain the color for pixels from the known (input) region while hallucinating the rest with a plausible scene completion (typically extrapolation). Guided by the binary mask in the input, we compute an $L1$ loss between the known pixels in the input and corresponding predicted colors. For the completed output, we apply a multi-scale $L2$ loss. This allows the model to coarsely regress the light direction and color, however it does not allow for generation of sharp features for the estimated light sources (as seen in Fig.~\ref{teaser-fig}(b) and Fig.~\ref{modelarch}). To obtain a high contrast HDR result that generates correct shadows $L2$ loss is not sufficient. 
 An ideal solution would be to use a ray tracer to render shadows and penalize the difference between the rendered images using ground truth and predicted  maps. This is non-trivial  and expensive for end-to-end training.
 
Considering only shadow casting, the intensity of a pixel in the shadow  is dependent on the integral of the environment map except directions blocked by the object. Thus, any pixel on the shadow plane can be approximated by the integral over the masked environment map.  We take inspiration from our intuition above, and Wasserstein distance, and introduce our novel ProjectionLoss. We select a set of randomized binary masks $P$ having the same size as our environment map, and create a one-dimensional vector of length $|P|$ (the number of images in the set) by integrating the pixel intensities in corresponding masked images. On a 0-valued background, the masks contain polygons generated with randomization whose height and width range from $10\%$ to $40\%$ of the corresponding image dimensions and filled with value of $1$. For further illustration of ProjectionLoss, and examples of the masks used, see Appendix~\ref{sec_proj} in supplementary materials. The final loss, termed \textit{ProjectionLoss}, is the  $L1$ distance between the vectors corresponding to predicted and ground truth environment maps. We show this error using $|P|=50$ projection masks in Fig.~\ref{angularerrorsamples}(a) for each environment map shown in row (c), with the first image being the reference. In Fig.~\ref{angularerrorsamples}(d) we show rendered spheres using the corresponding environment maps. We can see that lower values of ProjectionLoss correlate with both lower angular error and better visual match of lighting direction to the reference. This is also supported in our ablation study results in Sec.~\ref{sec_quantitative}. 

For a ground truth HDR image $G$, a mask indicating unknown pixels $M$, a predicted image $I$, and a set of projection masks $P$, the final image-based loss is defined as: 
\begin{equation} \label{eq_img_loss}
\begin{split}
Loss_{projection}&= | \forall_{P_i \in P}(\sum{(I*P_i)}-\sum{(G*P_i)}) |_1\\
Loss_{image}&=w_1|I*M-G*M|_1  + w_2 || I-G||_2\\
& + Loss_{projection}\\
\end{split}
\end{equation}
ProjectionLoss is related to diffuse convolution or cosine loss used in previous works \cite{gardner-sigasia-17,song2019neural}. While filtered, or down sampled integral can represent diffuse or low frequency lighting information, we find that our formulation is able to capture high frequency lighting better. 
To further understand the value of ProjectionLoss for lighting estimation, and to compare with other measures, such as SSIM \cite{ssim} and Mean Squared Error (MSE), we performed a detailed experiment with user study as described in Appendix~\ref{sec_proj}. We first confirmed that SSIM on rendered images is a good baseline for retrieval of images with similar lighting on objects. We then correlated retrieval of similar environment maps using ProjectionLoss, SSIM and MSE. Quantitatively, we found the intersection of top-5 retrievals by SSIM (on the rendered images) and those using ProjectionLoss (on the environment map) to be $1.6\pm0.7$, while it was $0.6\pm0.5$ using MSE (on the environment map).
Based on the above we believe that our proposed ProjectionLoss effectively trains the model for light estimation, such that the end result for rendering is accurate with respect to ground truth. We also show that MSE on the environment map is insufficient for training accurate light estimation. Its use as a metric of comparison, as done in previous works, would not correlate well with the final application.

\subsection{Adversarial loss and ClusterLoss}\label{sec_advloss}
To generate perceptually pleasing completions to the scene, we train the model using a GAN loss. Adversarial training was proposed in ~\cite{GoodfellowGAN2014}, where a discriminator was trained to classify samples from the training set as ``real'' and those produced by a generator as ``fake''. The generator has a loss component for classification of generated samples as ``real''. The  networks are trained by alternating the update of their weights.  Since their advent, there have been various improvements on GANs for training stability and result quality. As discussed previously, many of these have focused on single objects such as faces \cite{karras2018progressive} or increased stability using extremely large datasets or architectures \cite{biggan}.  However, we found that training a mobile-friendly GAN architecture with fewer than $3,000$ images with high variability in content resulted in mode collapse or non convergence to plausible images. 

It has been shown that providing an additional task to the discriminator using conditional labels or self-supervision can increase stability \cite{pmlr-v70-odena17a}. Assigning semantic labels (e.g, room types) is  non-trivial for panoramic images since the viewpoint could be from between two rooms.  Recent work using self-supervision with rotation  prediction \cite{Chen2018SelfSupervisedGV} has been promising. However, for equirectangular images the  possible rotations would be flips along the image axis, of which that along the vertical image axis is valid for a scene.

We thus propose a novel scheme combining the learnings from above methods. We begin with the recognition that our application needs to generate ``plausibly similar looking images'' as the training set, hence the idea to use appearance features to form a secondary task. For the appearance features we extract equal sized patches on a grid from each training image, and calculate the ORB~\cite{ORB} descriptor and mean color for each patch. Using them as features we assign a K-means-derived cluster ID to each image (we used $K=5$).
 Along with the typical real vs. fake classification, the discriminator is supervised to classify the K-means assigned cluster ID for each real image with the proposed ClusterLoss. This allows us to train without additional heuristics or tricks, and avoid mode collapse, as the encoded latents are forced to spread out over the appearance clusters. Intuitively it also helps the discriminator focus on a low ambiguity appearance-based task that pays attention to spatial locations and details, thus avoiding cases where early on in the training process a single patch artifact can easily clue the discriminator that the image is fake (``easy wins'').  The discriminator is composed of residual blocks as detailed in Appendix ~\ref{sec_network_details}. We use softmax cross entropy loss for the adversarial training of our model and discriminator $D$.  With $y,p$ indicating the ground truth and predicted one-hot encoded vectors for the cluster classification respectively, our adversarial losses are:
\begin{equation} \label{eq_adv_loss}
\begin{split}
Loss_{cluster}&= \sum_{k=1}^{5}{y_{o,k} \log(p_{o,k})}\\
Loss_{fakereal}&= \log(D(G))+\log(1-D(I))\\
Loss_{adversarial}&=-\log(D(I))  \\
\end{split}
\end{equation}

\subsection{Implementation and adversarial training}
During training we mask the ground truth images using randomly generated polygonal regions, and provide them as corresponding incomplete input environment map. The regions are of irregular shape and can be non-contiguous to generalize our model to complete any partial input. 
For input, we tonemap the ground truth HDR environment map by dividing the pixels by the average intensity (exposure compensation), applying a gamma with  value of $2.2$, and  final normalization and clipping of the intensity to $[-1.0,1.0]$. The color channels in the unknown pixels are filled with numbers from uniform random distribution in $[-1.0,1.0]$ range.  We use the logarithmic transform from Sec.~\ref{sec_data} for the ground truth used in evaluation of losses. The  components of $Loss_{image}$ are weighted using $w_1=0.5$, $w_2=0.01$.  For all image-based losses, we also weight the pixels to compensate for the  subtended angle on the sphere.
\begin{equation}\label{eq_finalloss}
	\begin{split}
		Loss_{EnvMapNet}&=  Loss_{image} + \\
		& Loss_{adversarial} + Loss_{cluster}\\
		Loss_{discriminator}&= Loss_{fakereal} +  Loss_{cluster} \\
	\end{split}
\end{equation}

We train the networks  by minimizing $Loss_{EnvMapNet}$  and $Loss_{discriminator}$  respectively.  We use the Adam optimizer, a batch size of 16, and the learning rate starting at $0.0002$, and decayed by half every $50$ epochs after the first $100$ epochs. We alternate between training the discriminator and EnvMapNet after each mini-batch.


\begin{figure*}[!h]
	\centering
	\begin{tabular}{ccccc}
		\shortstack{%
			\includegraphics[align = b, width = 0.33\linewidth]{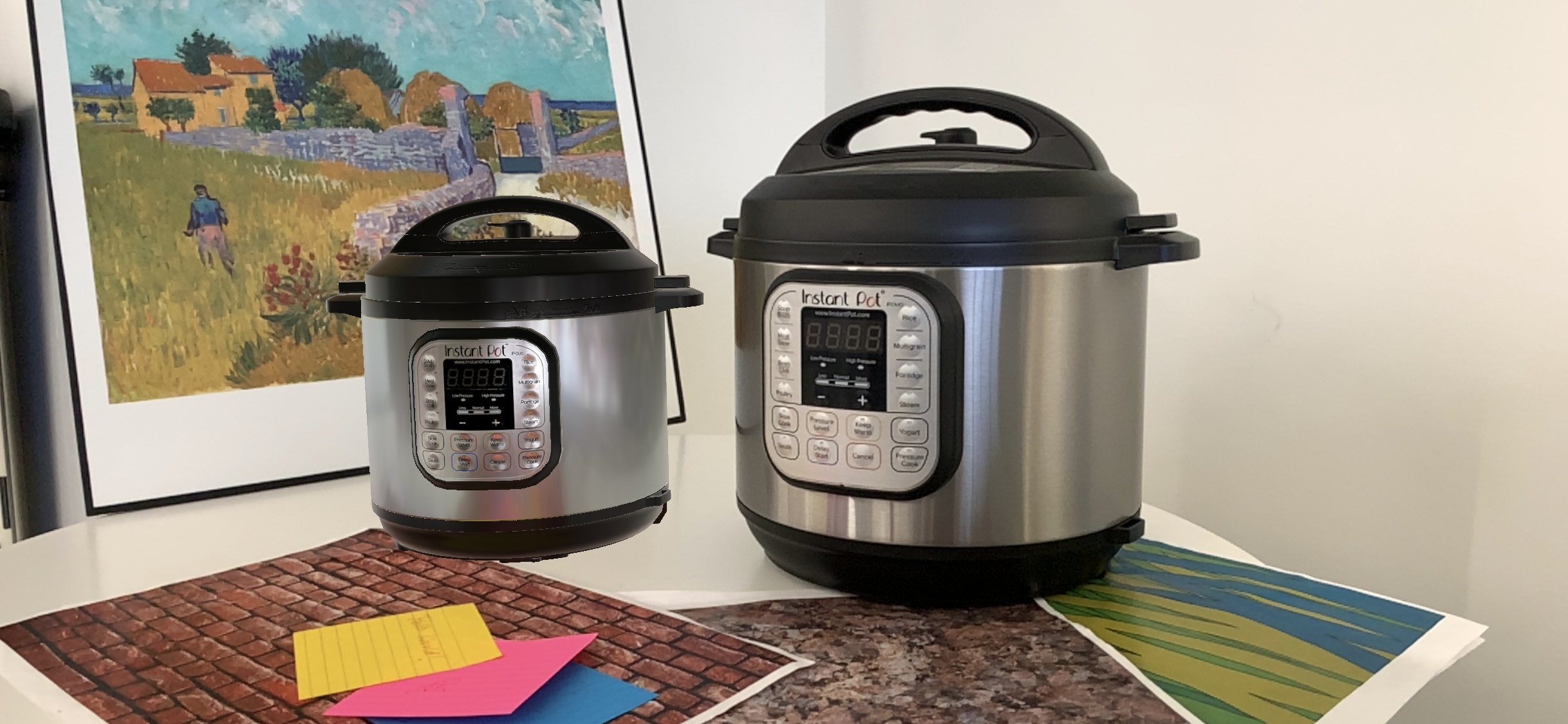} \\ (a) \\
			\includegraphics[align =b, width = 0.33\linewidth]{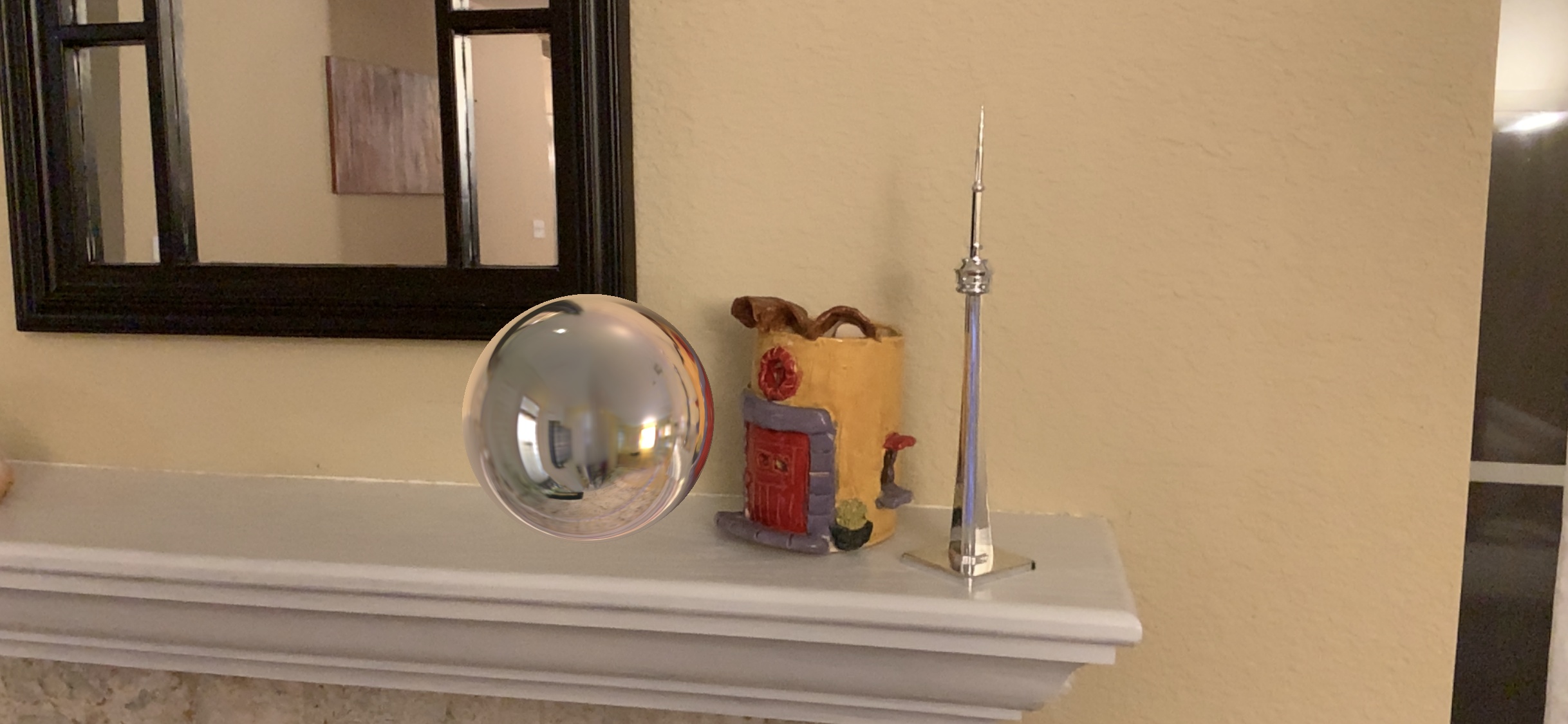}
		} &
		
\includegraphics[align = b, width = 0.155\linewidth]{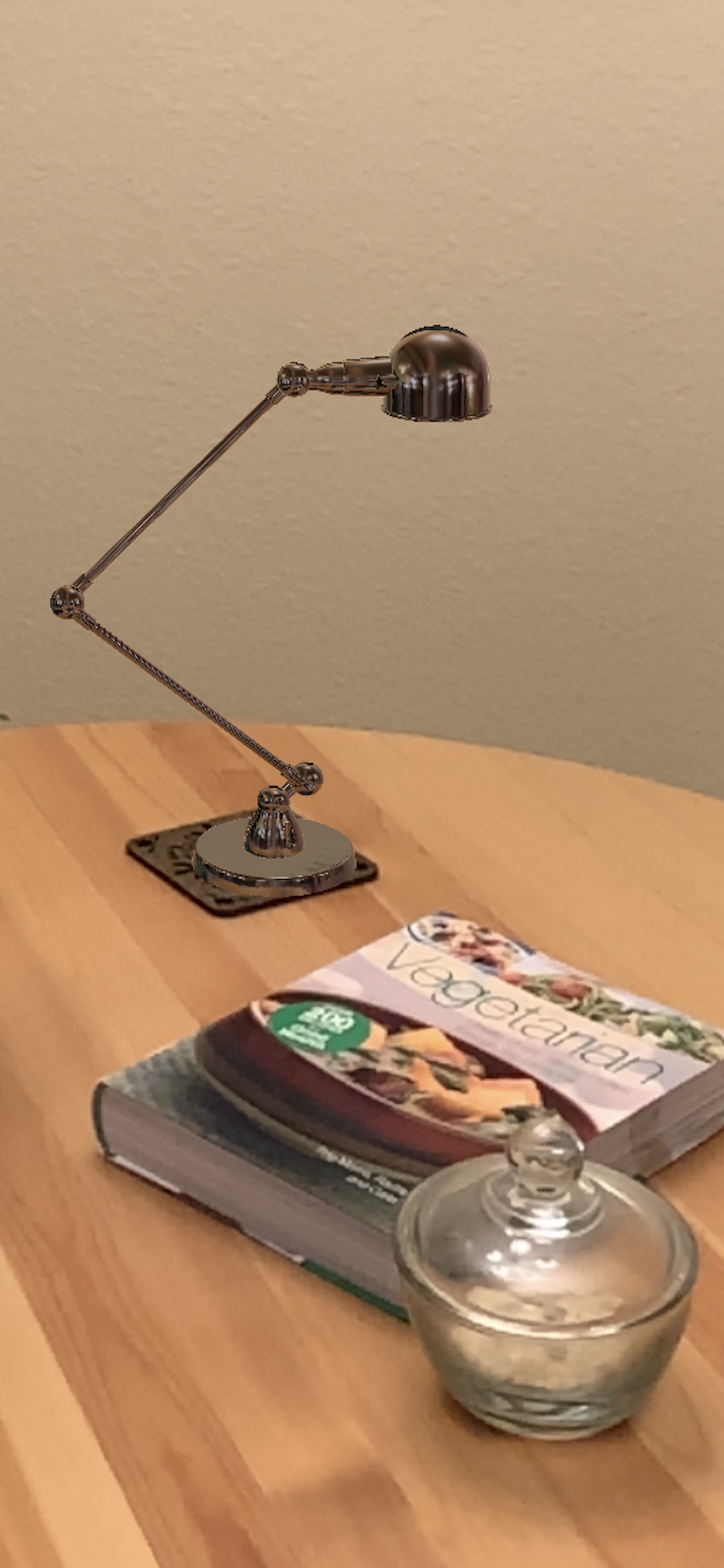} &
		\includegraphics[align = b, width = 0.155\linewidth]{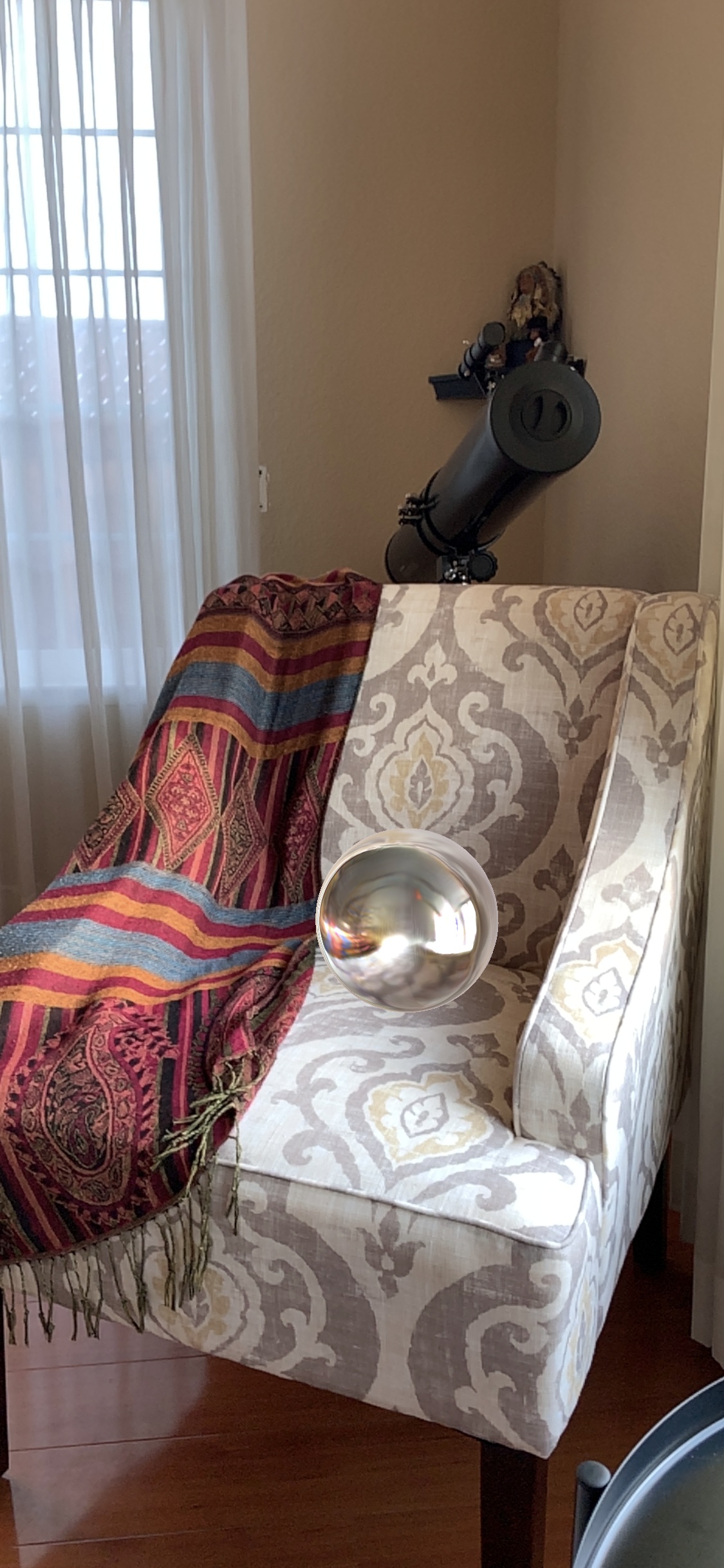}		&
		\includegraphics[align = b, width = 0.155\linewidth]{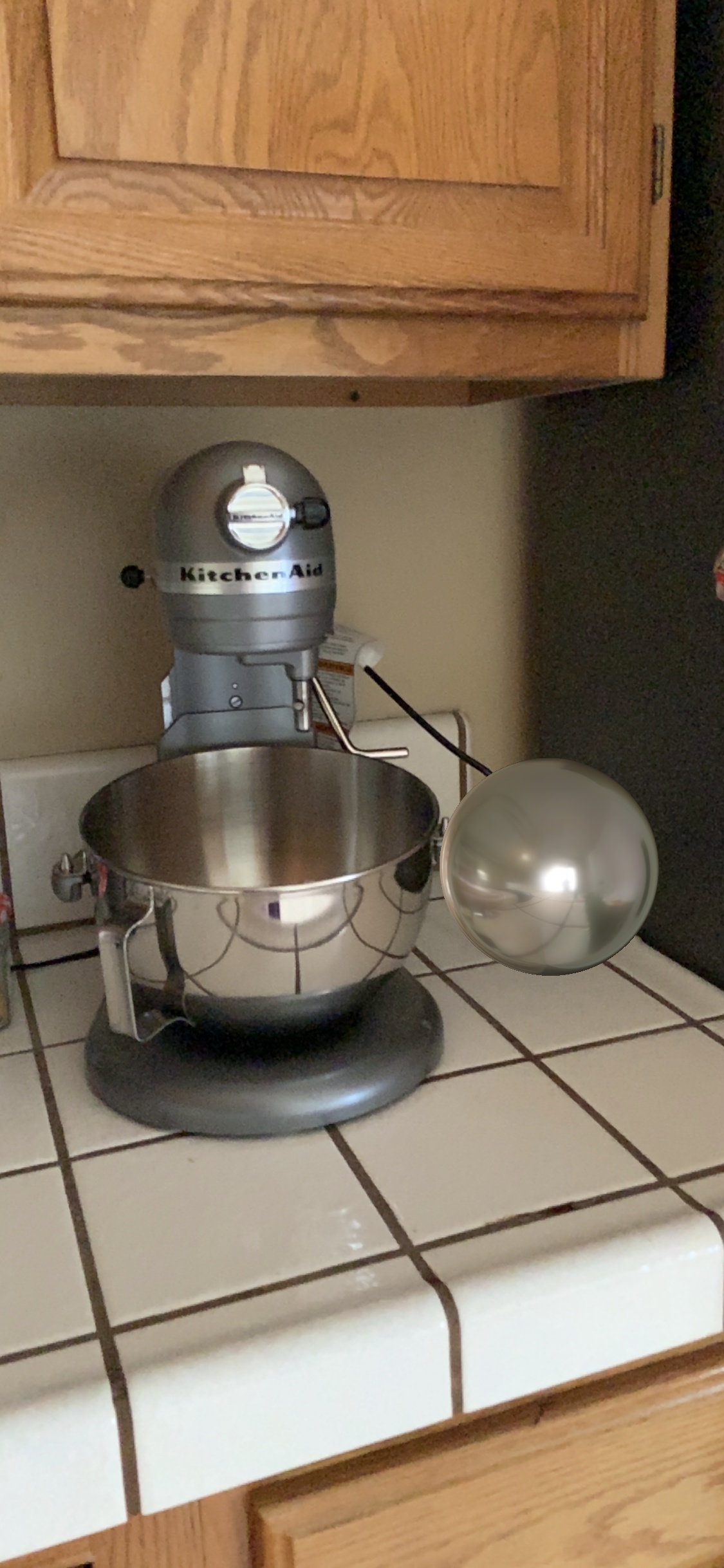}	\\

		(b)	& (c) & (d) & (e) \\

	\end{tabular}
	\caption{Snapshots from our real-time mobile app. (a) A real and virtual rendering of same object, we can see that the lighting direction and color matches closely. (b) Lighting on the virtual mirror finish  sphere is coherent with other real objects in the scene and the reflection formed is also a plausible completion to the scene. (c) A virtual lamp on the table, with its top specular highlight matching the direction  on the glass object on the lower right. (d)-(e) Virtual metallic spheres with different roughness. See videos in \textbf{supplementary material}.}\label{fig_apprender}
\end{figure*}

\section{Metrics for benchmarking}
\label{sec_metrics}
Visual coherence in AR aims to give a user the illusion that an inserted virtual object belongs in the real scene. However algorithm development, especially in the domain of deep learning, involves many heuristics and parameter tuning. We therefore believe that establishing quantitative metrics, separate from loss functions, that  correlate with the target application setting, is important to easily compare different methods. Several previous methods have used MSE on environment maps in RGB, intensity or log formulation as a metric. However, as discussed in Sec.~\ref{sec_regloss}, it is not well suited to measure accuracy such that the final rendering matches ground truth. 

At the core of it we want to solve for two aspects of the environment map: lighting accuracy and perceptual plausibility. Real-time mobile renderers use analytical (point) lights to cast shadows. Hence lighting accuracy is correlated to direction of top-n dominant light sources. Since we cannot (and should not) generate the exact unseen test room for reflection, measures such as MSE or SSIM  \cite{ssim,msssim} for one-to-one image comparisons are unsuitable. Instead we need to use plausibility of the hallucinated scene as a metric.  We start with a discussion on lighting accuracy metrics followed by quantitative measures for perceptual quality. Our goal is to define metrics that can be used by any future work to objectively compare techniques. Thus we establish a repeatable benchmark that uses the publicly available Laval HDR dataset \cite{gardner-sigasia-17}, and reference implementations of the metrics that we will provide.

Methods that can extract light sources from environment maps include median cut~\cite{Debevec:2006:MCA:1185657.1185688}, variance minimization~\cite{Viriyothai:2009:VML:1599301.1599393}, and parametric fitting from peak finding \cite{Gardner_2019_ICCV}. In the case of mobile AR, real-time renderers need fast decomposition techniques, hence we use the latter method to find centers and extent of light sources. The angular coordinates of the ellipse center are used to measure the directional accuracy with respect to ground truth. For a given HDR environment map, we iteratively find a seed pixel with maximum intensity, and grow the region connected to the seed till we reach $30\%$ of the peak value. This is repeated until the peak value found is less than $90\%$ of the largest. For each such region, we fit an ellipse covering its convex hull. Example images with parametric light fitting are shown in Fig.~\ref{angularerrorsamples}(c).

To measure the \textit{AngularError} between the ground truth and estimated environment maps, we extract (at most) five parametric lights from each. For each ground truth light we find the minimal angular error to the predicted light set, and vice-versa. We take the mean over the errors from both ground truth lights and predicted lights, as the final AngularError between the two environment maps. Fig.~\ref{angularerrorsamples} demonstrates the correlation of the AngularError metric, with examples of low and high errors with respect to a selected reference. 
We can see that environment maps with lower errors produce similar lighting directions on the reference sphere, compared to those with higher errors, thus making it a suitable metric for comparison.

To measure perceptual plausibility, user studies are commonly used for their insight into final experience quality. However, user studies are hard to repeat or compare across publications. We take inspiration from work in GANs, that measure quality of synthesized images using Frechet Inception Distance (FID)~\cite{Heusel:2017:GTT:3295222.3295408}. The measure is defined to evaluate similarity of distributions between two image sets, in contrast to perceptual metrics such as SSIM that require a reference image for each given test image. That is, the FID is low between two sets of images that have similar image features, and overall diversity among the different generated samples is the same as in the reference samples. This metric is thus good  to demonstrate that not only do the details in any generated environment map match realistic patches from training images, but the set of predicted images also have variety in content without over-fitting. 

\section{Results}\label{sec_results}
To show the effectiveness of our method for rendering virtual objects in AR, we start with qualitative results from real-world applications of our model, followed by comparisons to previous work and quantitative benchmark results.

 \subsection{Qualitative results}\label{qualitative_results}
In Fig.~\ref{fig_apprender} we show results from our prototype iOS app used in real-world scenes. We use device pose and plane geometry provided by ARKit \cite{arkit}  to warp the input camera image into a partial environment map. The completed environment map from running our model on the device is then used with SceneKit\footnote{https://developer.apple.com/documentation/scenekit/} to render the virtual object\footnote{3D models from https://developer.apple.com/augmented-reality/quick-look/}. The inference time is under $9$ ms on an iPhone~XS, allowing updates at high frame rates depending on desired user experience. Please see the \textbf{supplementary material for videos} from our application, where we use an update rate of $10$fps for our results.  Fig.~\ref{fig_apprender}(a) shows a real cooker in the right, and a rendered version on the left. Note that  the material properties of the brushed steel body do not match exactly. We can clearly see that the lighting direction closely matches the reference real-world object, providing the correct specular highlights. The accurate lighting, combined with perceptually plausible reflection clearly makes the rendering an immersive AR experience. In Fig.~\ref{fig_apprender}(b) we render a mirror sphere into a scene to show the impact of generating plausible reflections. Though the input observes a very small part of the environment, we are able to create a believable AR experience by completing the environment map with a detailed scene. Even though it cannot match the actual room in every detail, the perceptual impact and coherence of the virtual object with other objects, such as the mirror on the wall, and metallic tower on the right, is evident. In Fig.~\ref{fig_apprender}(c)-(e) we provide more examples of inserted virtual objects that have  lighting that is coherent with the  real objects in the field of view.

From the images in Fig.~\ref{teaser-fig} and ~\ref{fig_apprender} we highlight a few aspects of our method. For each of the scenes, that vary in time of day and FoV, not only is the light direction coherent with the scene but so is the type/shape/size of light source generated by our method. Fig.~\ref{teaser-fig}(e) shows a virtual teapot on a table illuminated with an unseen artificial light. The roughness of the teapot was $0.2$, and metalness was $1.0$ in SceneKit. The light direction and highlights from our estimate closely match those  on the real teapot (blue) and sphere (pool ball) in the scene. In Fig.~\ref{fig_apprender}(b) and (d) the light sources have appearances closer to that of a window/door that are common and plausible in those environments.

We show qualitative comparisons to recent work from Srinivasan~\etal \cite{srinivasan20lighthouse}, and their re-implementations of Neural Illumination~\cite{song2019neural} and Deep Light~\cite{LeGendre:2019:DLI:3306307.3328173} in \textbf{Appendix~\ref{app_compare}}.

%

\begin{table}[!t]
	\centering
	\begin{tabular}{|c|c|c|}
		\hline
		\textbf{Method} & \textbf{FID} & \textbf{AngularError} \\ \hline
		{\shortstack{EnvMapNet  (Ours)}} & 52.7 & 34.3$\pm$18.5\\   \hline
		{\shortstack{Ours without  ProjectionLoss}} &  77.7 & 39.2$\pm$29.9\\ \hline
		{\shortstack{Ours without  ClusterLoss}} &  203 & 75.1$\pm$25\\ \hline
		{\shortstack{Gardner~\etal   \cite{gardner-sigasia-17}}} & 197.4 & 65.3$\pm$24.5\\ \hline
		{\shortstack{Artist IBL (Fixed)}} & - & 46.5$\pm$15.4\\
		
		\hline

	\end{tabular}
	\caption{Quantitative comparisons and ablation studies (Sec.~\ref{sec_quantitative}).} \label{tab_metrics}
\end{table}

\subsection{Quantitative evaluation and benchmarking} \label{sec_quantitative}
\begin{figure}
	\centering
	\begin{tabular}{@{}c@{ }c@{ }c@{ }c@{ }c@{ }c@{}}
		\shortstack{Input\\Crop} &	\shortstack{EnvMapNet \\(Ours)} & \shortstack{ Gardner~\etal \\ \cite{gardner-sigasia-17}}  & \multicolumn{3}{c}{\shortstack{Rendered objects\\Ours - GT - ~\cite{gardner-sigasia-17} }}\\
		\includegraphics[align = c, width = 0.11\linewidth]{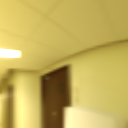} &
		\includegraphics[align = c, width = 0.24\linewidth]{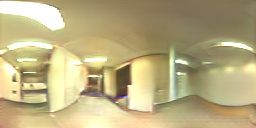} &
		\includegraphics[align = c, width = 0.24\linewidth]{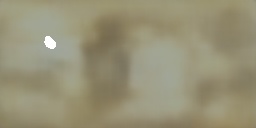} &
		\includegraphics[align = c, width = 0.11\linewidth]{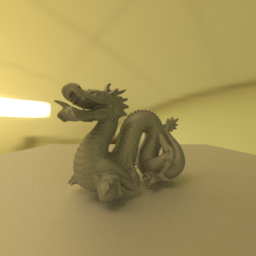} &
		\includegraphics[align = c, width = 0.11\linewidth]{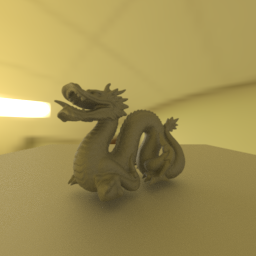} &
		\includegraphics[align = c, width = 0.11\linewidth]{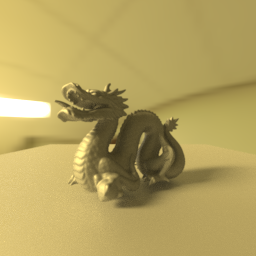} \\
			\includegraphics[align = c, width = 0.11\linewidth]{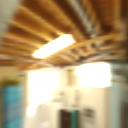} &
		\includegraphics[align = c, width = 0.24\linewidth]{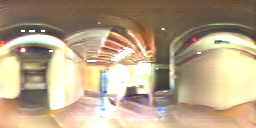} &
		\includegraphics[align = c, width = 0.24\linewidth]{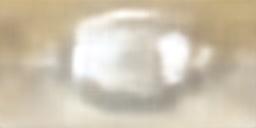} &
		\includegraphics[align = c, width = 0.11\linewidth]{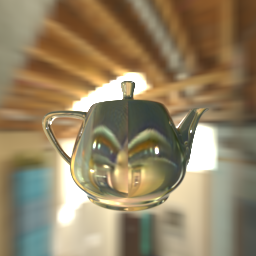} &
		\includegraphics[align = c, width = 0.11\linewidth]{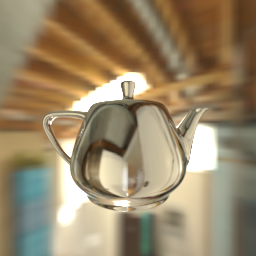} &
		\includegraphics[align = c, width = 0.11\linewidth]{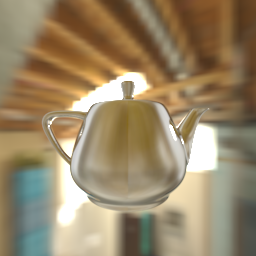} \\
		\includegraphics[align = c, width = 0.11\linewidth]{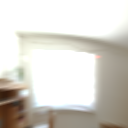} &
		\includegraphics[align = c, width = 0.24\linewidth]{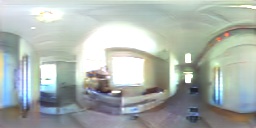} &
		\includegraphics[align = c, width = 0.24\linewidth]{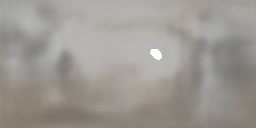} &
		\includegraphics[align = c, width = 0.11\linewidth]{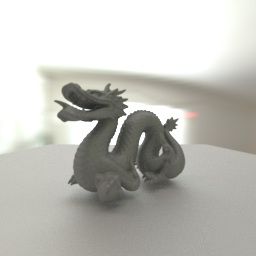} &
		\includegraphics[align = c, width = 0.11\linewidth]{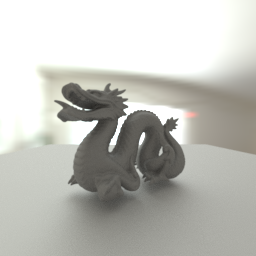} &
		\includegraphics[align = c, width = 0.11\linewidth]{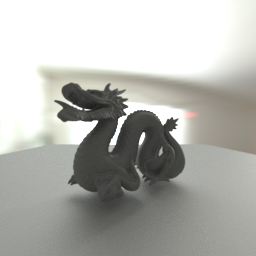} \\
		\includegraphics[align = c, width = 0.11\linewidth]{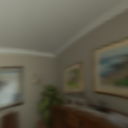} &
		\includegraphics[align = c, width = 0.24\linewidth]{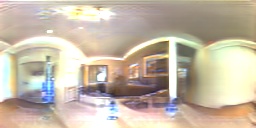} &
		\includegraphics[align = c, width = 0.24\linewidth]{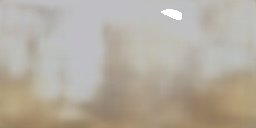} &
		\includegraphics[align = c, width = 0.11\linewidth]{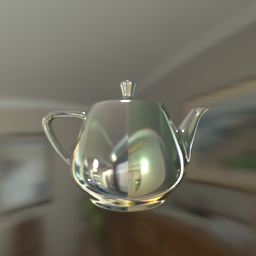} &
		\includegraphics[align = c, width = 0.11\linewidth]{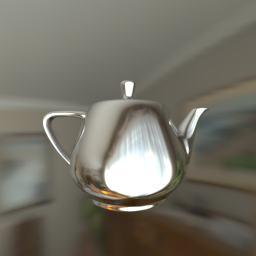} &
		\includegraphics[align = c, width = 0.11\linewidth]{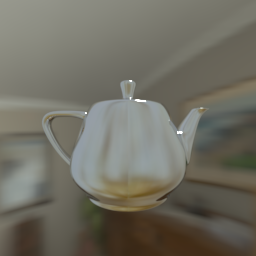} \\
		\includegraphics[align = c, width = 0.11\linewidth]{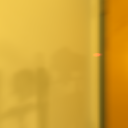} &
		\includegraphics[align = c, width = 0.24\linewidth]{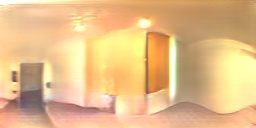} &
		\includegraphics[align = c, width = 0.24\linewidth]{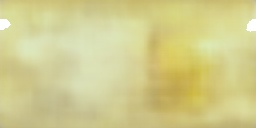} &
		\includegraphics[align = c, width = 0.11\linewidth]{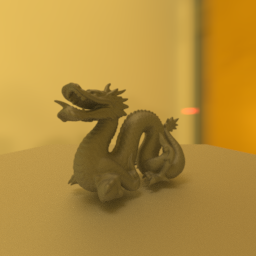} &
		\includegraphics[align = c, width = 0.11\linewidth]{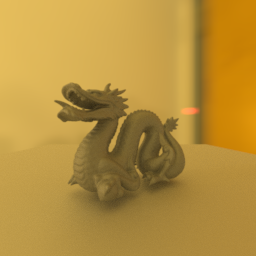} &
		\includegraphics[align = c, width = 0.11\linewidth]{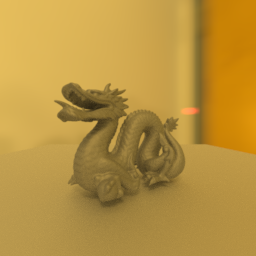} \\

	\end{tabular}
	\caption{Sample benchmarking results (Sec.~\ref{sec_quantitative} and Appendix~\ref{sec_appendix_res}). }
	\label{fig_selected_res_main}
\end{figure}

For quantitative benchmarking we use the metrics detailed in Sec.~\ref{sec_metrics} on the publicly available Laval dataset~\cite{gardner-sigasia-17}, and show the results in Table~\ref{tab_metrics}. We show some example results in Fig.~\ref{fig_selected_res_main} with rendered objects using results and ground truth (GT)  environment maps. All images are best seen on a  color monitor with magnification. We highly encourage the reader to see \textbf{Appendix~\ref{sec_appendix_res}} for details and more results with renderings.  

As discussed in Sec.~\ref{sec:related}, though several methods have focused on light estimation from single images, only a few generate HDR RGB environment maps. Recent techniques that estimate lighting such as ~\cite{song2019neural} and ~\cite{LeGendre:2019:DLI:3306307.3328173}, have not shared code, model weights nor use  public datasets. There is also a lack of a standard set of metrics for comparison. The method by Gardner~\etal~\cite{gardner-sigasia-17} is the current state-of-the-art method that is most comparable to our method, uses the same public dataset, and is compared in the above recent works as well. They have provided a web interface to obtain results from their method\footnote{http://rachmaninoff.gel.ulaval.ca:8001/}. Since we  use the same splits for train/test, we provide detailed quantitative and qualitative comparisons with their results. 

We summarize the quantitative metrics in Table~\ref{tab_metrics}, calculated over the $250$ test images. The results from using our  method are reported as \textbf{EnvMapNet}. We introduced two novel loss functions in our method, for which we also conducted ablation studies.  For \textbf{Ours without ProjectionLoss} we observe higher angular error than our full model EnvMapNet, as expected from our discussions in Sec.~\ref{sec_regloss}. When we trained \textbf{Ours without ClusterLoss}, for the discriminator, the model  often suffered from instability as discussed in Sec.~\ref{sec_advloss}, and could not generate  textures with fine details, hence the higher FID and angular error. 

The errors for Gardner~\etal~\cite{gardner-sigasia-17} are higher than our complete pipeline. As shown with example images (Fig.~\ref{fig_selected_res_main} and Appendix~\ref{sec_appendix_res}), the environment map generated by their method lacks fine details for the scenes hence the higher FID score. The angular error is also higher on average, which could be explained from our experiments in Sec.~\ref{sec_regloss} that show MSE loss is insufficient to train for accurate lighting. Both these aspects can be clearly observed to affect the use of the environment maps in rendering objects with both mirror and diffuse finish, along with the cast shadows.

As a baseline measurement, we consider that AR applications, like games, may use a fixed environment map, created by an artist, for lighting. We obtained such an environment map designed for indoor scenes as detailed in Appendix~\ref{app_artist_ibl}. We calculate  the angular error metric with respect to this artist designed environment map, and report it in Table~\ref{tab_metrics} as \textbf{Artist IBL}. 

As demonstrated through the various qualitative examples and quantitative benchmarking, our method, with lowest FID and AngularError, can produce high quality environment maps both for visually pleasing reflections, and accurate lighting of the virtual objects.


\section{Conclusions}
We presented the first method that, given a small field-of-view LDR camera image, can estimate a high resolution HDR environment map in real-time on a mobile device. The result can be used to light objects of any material finish (mirror to diffuse) for augmented reality. We made two novel contributions in the training of our neural network, with ProjectionLoss for the environment map, and ClusterLoss for the adversarial loss. This enabled our method to synthesize environment maps with accurate lighting and perceptually plausible reflections.  We proposed two metrics to measure both these aspects of the estimated environment map. Through qualitative and quantitative comparisons we demonstrated that our method  reduces the angular error in parametric light direction by more than $50\%$, along with a $3.7$ times reduction in FID. We showcased a real-world mobile application that is able to run our model in real-time (under $9$ms) and render visually coherent virtual objects in novel environments.

{\small
	\bibliographystyle{ieee_fullname}
\bibliography{EnvMapNet_bib}
}

\makeatletter\@input{supplementaryaux.tex}\makeatother

\end{document}


\title{HDR Environment Map Estimation for Real-Time Augmented Reality \\Supplementary Appendix }
\author{Gowri Somanath\\
	Apple\\
	{\tt\small gowri@apple.com}
	\and
	Daniel Kurz\\
	Apple\\
	{\tt\small daniel\textunderscore kurz@apple.com}
}
\maketitle

\section{Video of mobile application} \label{app_video}
A video providing an overview of our method and a demonstration of our mobile application in the real world using iPhone~XS is available at \url{https://docs-assets.developer.apple.com/ml-research/papers/hdr-environment-map.mp4}

\begin{figure}[!t]
	\centering
	\includegraphics[align=c,width=0.99\linewidth]{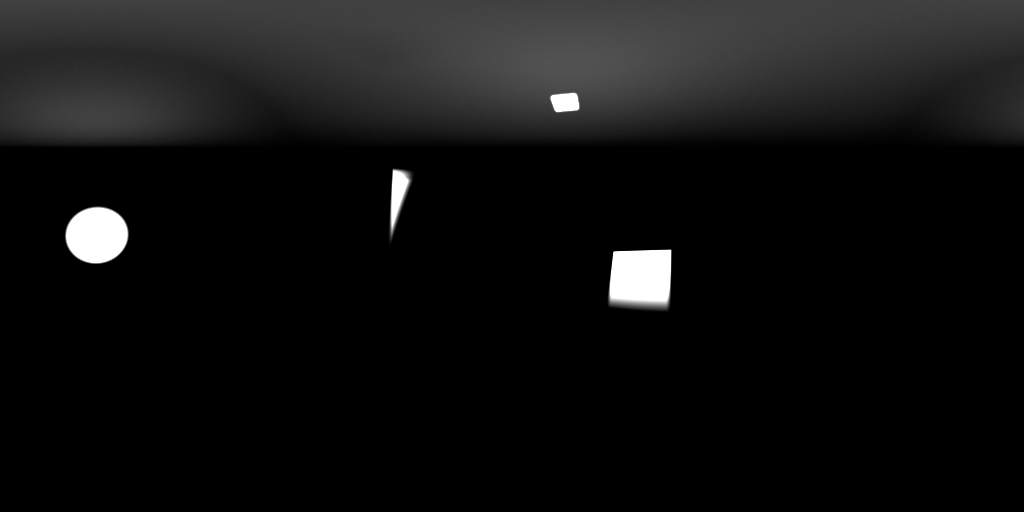}
	\caption{Artist-created IBL used for baseline measurement.} \label{fig_artistibl}
\end{figure}

\section{Artist-created environment map} \label{app_artist_ibl}
In Figure~\ref{fig_artistibl} we show the artist-created environment map used as a baseline in our benchmarking. The artist designed it to satisfy aesthetic and lighting requirements. For lighting, the intensities are selected to make sure the objects were well exposed and that middle gray is retained. The aesthetic brief was to have a ``studio lighting'' feel with a broad area light from behind the camera and from above.

\begin{table}
	\begin{tabular}{|l|}
		\hline
		\textbf{EnvMapNet-conv-block} \\
		\hline
		short-cut,x=input \\
		Repeat 5 times: \\
		\hspace{10mm} x=BatchNormalization(x)  \\
		\hspace{10mm} x=LeakyReLu(slope=0.2)(x) \\
		\hspace{10mm} x=Convolution(kernel=(3,3),filters=16)(x) \\
		\hspace{10mm} x=Concatenate(x,short-cut)\\
		\hspace{10mm} short-cut=x\\
		output=x \\
		\hline
		\hline
		\textbf{EnvMapNet-downsample-block} \\
		\hline
		x=input \\
		x=Convolution(kernel=(3,3),filters=dk)(x) \\
		x=AveragePool2D(x)\\
		output=x \\
		\hline
		\hline
		\textbf{EnvMapNet-upsample-block} \\
		\hline
		x=input \\
		x=NearestNeighbourUpsample2x(x)\\
		x=Convolution(kernel=(3,3),filters=uk)(x) \\
		output=x \\
		\hline
		\hline
		\textbf{Discriminator-residual-block} \\
		\hline
		sc=AveragePool2D(input)\\
		sc=Convolution(kernel=(3,3),filters=ak)(sc)\\
		x=input\\
		Repeat 2 times: \\
		\hspace{10mm}x=BatchNormalization(x)  \\
		\hspace{10mm}x=LeakyReLu(slope=0.2)(x) \\
		\hspace{10mm}x=Convolution(kernel=(3,3),filters=ak)(x) \\
		x=AveragePool2D(x) \\
		output=Add(x,sc)\\
		\hline
	\end{tabular}
	\caption{Building blocks used in EnvMapNet and discriminator.}
	\label{tab_cnnblocks}	\label{tab_discblock}
\end{table}

\section{Network details}
\label{sec_network_details}
In Table~\ref{tab_discblock} we provide the building blocks of the model architectures used in our method. Spectral Normalization ~\cite{miyato2018spectral} is used in all  the convolutional layers.

Our proposed model, EnvMapNet, consists of an encoder and decoder as shown in paper Figure~\ref{modelarch}.  The encoder is composed of seven sets of \textit{EnvMapNet-conv-block} and \textit{EnvMapNet-downsample-block}, with dk=[$64$, $128$, $128$, $128$, $256$, $256$, $512$] for each consecutive block respectively. The resulting latent vector is convolved with a 1$\times$1 kernel to output $64$ filters. The decoder mirrors the encoder by using \textit{EnvMapNet-conv-block} and \textit{EnvMapNet-upsample-block}, with uk=[$512$,$256$,$256$,$128$,$128$,$128$,$64$].  The final output is produced by a  3$\times$3 convolution to produce $3$ channels for RGB, followed by a tanh activation. We use skip connections between same sized layers of the encoder and decoder. 

The discriminator is composed of residual blocks with ak=[$64$,  $128$, $256$, $256$,  $256$, $256$,  $256$] for consecutive blocks respectively. The  outputs from the discriminator are the binary classification of real or fake, and the classification into the K-means cluster ID. Each is obtained by convolution layer with the corresponding number of output channels and global average pooling.

\begin{figure}[!t]
	\centering
	\includegraphics[align=c,width=0.9\linewidth]{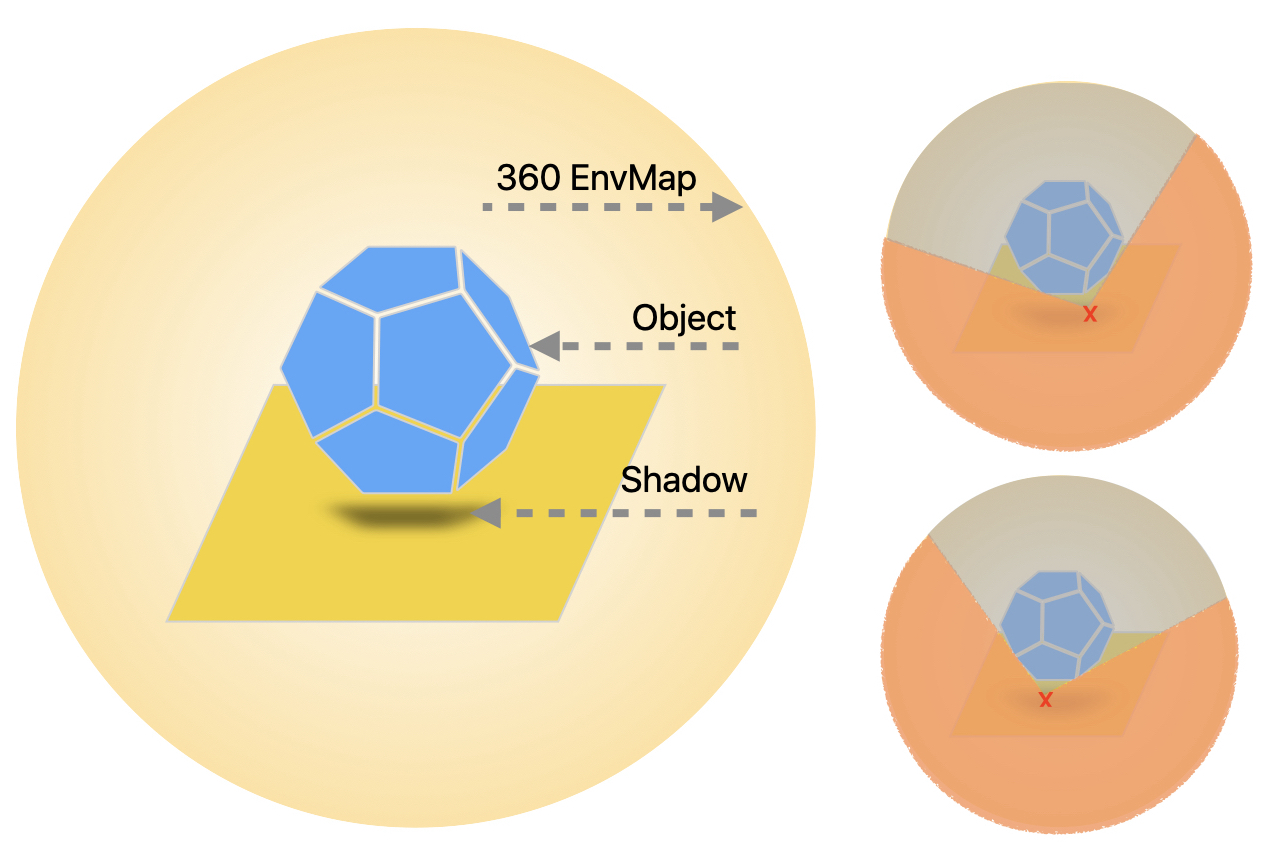} \hfill
	\caption{Intuitive understanding of ProjectionLoss and its relation to shadow casting. The loss is defined in Section~\ref{sec_regloss}.} \label{app_proj_shadow}
\end{figure}

\begin{figure}[!t]
	\centering
	\includegraphics[align=c,width=0.45\linewidth]{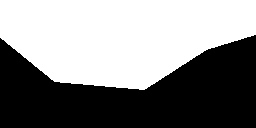} \hfill
	\includegraphics[align=c,width=0.45\linewidth]{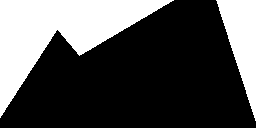} \\
	\includegraphics[align=c,width=0.45\linewidth]{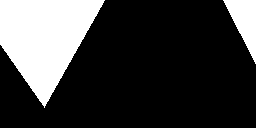} \hfill
	\includegraphics[align=c,width=0.45\linewidth]{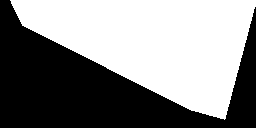}
	\caption{Example masks used for ProjectionLoss.} \label{app_fig3}
\end{figure}

\section{ProjectionLoss and user study details}\label{sec_proj}
In this section we expand on the background for ProjectionLoss, its relation to shadow casting, and the details of the user study and experiments discussed in paper Section~\ref{sec_regloss}.

\subsection{Relation to shadows}\label{app_proj_masks}
In Figure~\ref{app_proj_shadow} we provide an illustration to explain our motivation for ProjectionLoss. Consider an object (blue ball), lit by an environment map, and casting a shadow on the planar surface below. Accurate generation of shadows requires for every point on the plane (red cross) the computation of the integral over the part of the env map that is visible from that point (shaded orange), i.e. not occluded by the object (blue ball). To this end, the environment map is element-wise multiplied with a visibility mask. Our precomputed randomized masks (Figure~\ref{app_fig3}) used in projection loss are examples for such visibility masks for different points on the plane and different object shapes. As a result, ProjectionLoss encourages our predicted env maps to lead to shadows similar to ground truth.

\subsection{User study}
\begin{figure*}[!t]
	\centering
	\begin{tabular}{|l@{}c|ccccc|}
		\hline
		Metric & Reference Image & \multicolumn{5}{|c|}{Top-5 Retrievals}\\
		\hline
		(a) SSIM &\includegraphics[align=c,width=0.10\linewidth]{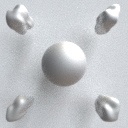} &
		\includegraphics[align=c,width=0.10\linewidth]{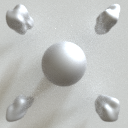} &
		\includegraphics[align=c,width=0.10\linewidth]{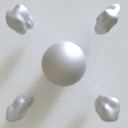} &
		\includegraphics[align=c,width=0.10\linewidth]{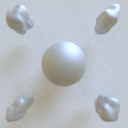} &
		\includegraphics[align=c,width=0.10\linewidth]{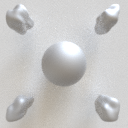} &
		\includegraphics[align=c,width=0.10\linewidth]{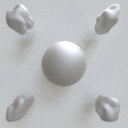} \\
		
		(b) ProjectionLoss &\includegraphics[align=c,width=0.10\linewidth]{images/shadows/9C4A8031_Panorama_hdr_normed_shadow.png} &
		\includegraphics[align=c,width=0.10\linewidth]{images/shadows/9C4A3419_Panorama_hdr_normed_shadow.png} &
		\includegraphics[align=c,width=0.10\linewidth]{images/shadows/9C4A0615_Panorama_hdr_normed_shadow.png} &
		\includegraphics[align=c,width=0.10\linewidth]{images/shadows/9C4A0162_Panorama_hdr_normed_shadow.png} &
		\includegraphics[align=c,width=0.10\linewidth]{images/shadows/9C4A0289_Panorama_hdr_normed_shadow.png} &
		\includegraphics[align=c,width=0.10\linewidth]{images/shadows/AG8A0319_Panorama_hdr_normed_shadow.png} \\
		
		(c) MSE &\includegraphics[align=c,width=0.10\linewidth]{images/shadows/9C4A8031_Panorama_hdr_normed_shadow.png} &
		\includegraphics[align=c,width=0.10\linewidth]{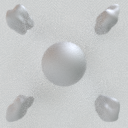} &
		\includegraphics[align=c,width=0.10\linewidth]{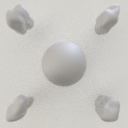} &
		\includegraphics[align=c,width=0.10\linewidth]{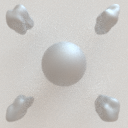} &
		\includegraphics[align=c,width=0.10\linewidth]{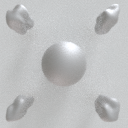} &
		\includegraphics[align=c,width=0.10\linewidth]{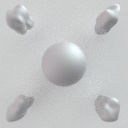} \\
		\hline
		\\
		\hline
		(a) SSIM &\includegraphics[align=c,width=0.10\linewidth]{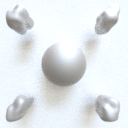} &
		\includegraphics[align=c,width=0.10\linewidth]{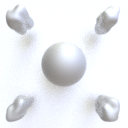} &
		\includegraphics[align=c,width=0.10\linewidth]{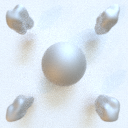} &
		\includegraphics[align=c,width=0.10\linewidth]{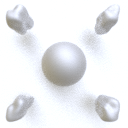} &
		\includegraphics[align=c,width=0.10\linewidth]{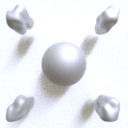} &
		\includegraphics[align=c,width=0.10\linewidth]{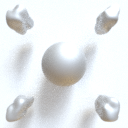} \\
		
		(b) ProjectionLoss &\includegraphics[align=c,width=0.10\linewidth]{images/shadows/9C4A0331_Panorama_hdr_normed_shadow.png} &
		\includegraphics[align=c,width=0.10\linewidth]{images/shadows/9C4A0484_Panorama_hdr_normed_shadow.png} &
		\includegraphics[align=c,width=0.10\linewidth]{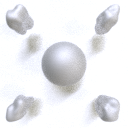} &
		\includegraphics[align=c,width=0.10\linewidth]{images/shadows/9C4A6981_Panorama_hdr_normed_shadow.png} &
		\includegraphics[align=c,width=0.10\linewidth]{images/shadows/AG8A3074_Panorama_hdr_normed_shadow.png} &
		\includegraphics[align=c,width=0.10\linewidth]{images/shadows/9C4A0380_Panorama_hdr_normed_shadow.png} \\
		
		(c) MSE &\includegraphics[align=c,width=0.10\linewidth]{images/shadows/9C4A0331_Panorama_hdr_normed_shadow.png} &
		\includegraphics[align=c,width=0.10\linewidth]{images/shadows/9C4A1287_Panorama_hdr_normed_shadow.png} &
		\includegraphics[align=c,width=0.10\linewidth]{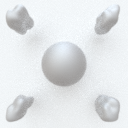} &
		\includegraphics[align=c,width=0.10\linewidth]{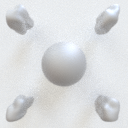} &
		\includegraphics[align=c,width=0.10\linewidth]{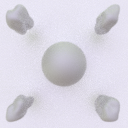} &
		\includegraphics[align=c,width=0.10\linewidth]{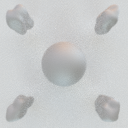} \\
		\hline
	\end{tabular}
	\caption{Retrieval for two reference images based on (a) SSIM on rendered images, (b) ProjectionLoss on equirectangular images, and (c) MSE on equirectangular images. A user study on $380$ reference images showed $95\%$ agreement with SSIM retrieval. Average intersection of participants’ selection of most similar rendered images with top-5 retrieval from ProjectionLoss was $1.6$, and from MSE was $0.6$.}
	\label{retrievalmetrics}
\end{figure*}

To further understand the value of ProjectionLoss for lighting estimation, and to compare to other measures, such as SSIM~\cite{ssim} and Mean Squared Error (MSE), we performed the following experiments. For each dataset environment map, we render an image of a scene with multiple geometric objects as shown in Fig.~\ref{retrievalmetrics}. 

First we establish a baseline metric to measure similarity between rendered images. We propose using the SSIM score, which is a metric often used to measure human perception and image similarity, to quantify the similarity of our rendered images. To validate this for our application, we perform a user study asking participants to choose from four provided options which rendered image looks most similar to a reference rendered image. The four options were randomly selected such that two of them were in the top-10 as retrieved by SSIM, and the other two from outside the top-10. Based on the results of $5$ study participants, each providing their selection for $380$ reference images, we found that on average human participants selected one of the top-10 SSIM images $95\%$ of the time. Hence retrieval of similar environment maps based on SSIM between rendered images is a baseline method to find environment maps that produce similar lighting on the objects.

We then evaluate the correlation between SSIM on rendered images with ProjectionLoss and MSE calculated on the corresponding (equirectangular) environment maps. We can observe that the retrieval of similar environment maps by ProjectionLoss is better matched with retrieval based on SSIM compared to using MSE, see Fig.~\ref{retrievalmetrics}. Quantitatively, we found the intersection of top-5 retrievals by SSIM (on the rendered images) and those using ProjectionLoss (on the environment map) to be $1.6\pm0.7$, while it was $0.6\pm0.5$ using MSE (on the environment map).

Based on the above we believe that our proposed ProjectionLoss is a good loss to train the network for estimating lighting such that the end result for rendering is accurate with respect to ground truth. Secondly, we show that MSE on the environment map is insufficient for training accurate light estimation, and its use as a metric of comparison or benchmarking, as done in previous works, would not correlate well with the final application.

\begin{figure*}
	\begin{tabular}{cccc}
		(a) Input &
		\includegraphics[align=c,width=0.22\textwidth]{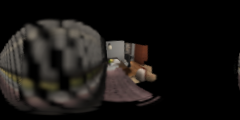} &
		
		\includegraphics[align=c,width=0.22\textwidth]{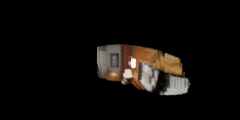} &
		\includegraphics[align=c,width=0.22\textwidth]{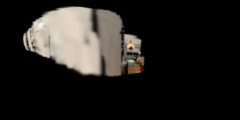} \\
		(b) Ground Truth &
		\includegraphics[align=c,width=0.22\textwidth]{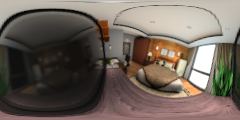} &
		\includegraphics[align=c,width=0.22\textwidth]{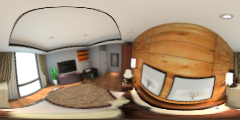} &
		\includegraphics[align=c,width=0.22\textwidth]{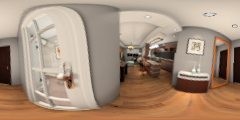} \\
		
		(c) Lighthouse \cite{srinivasan20lighthouse} &
		\includegraphics[align=c,width=0.22\textwidth]{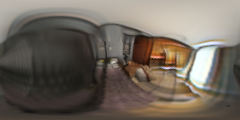} &
		\includegraphics[align=c,width=0.22\textwidth]{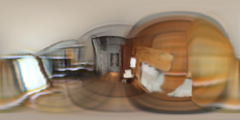} &
		\includegraphics[align=c,width=0.22\textwidth]{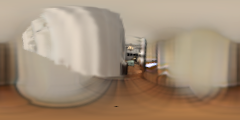} \\

		(d) Deep Light\cite{LeGendre:2019:DLI:3306307.3328173} &
		\includegraphics[align=c,width=0.22\textwidth]{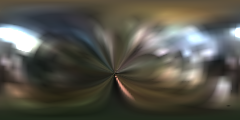} &
		\includegraphics[align=c,width=0.22\textwidth]{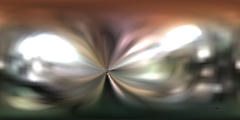} &
		\includegraphics[align=c,width=0.22\textwidth]{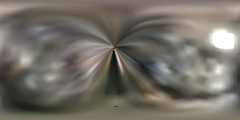} \\
		
		(e) Neural Illumination\cite{song2019neural} &
		\includegraphics[align=c,width=0.22\textwidth]{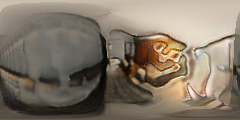} &
		\includegraphics[align=c,width=0.22\textwidth]{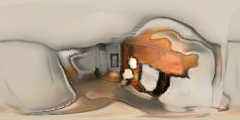} &
		\includegraphics[align=c,width=0.22\textwidth]{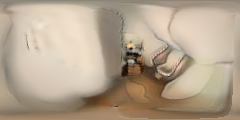} \\
		
		(f) EnvMapNet (Ours on (a)) &
		\includegraphics[align=c,width=0.22\textwidth]{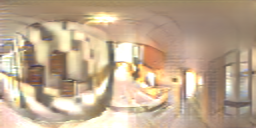} &
		\includegraphics[align=c,width=0.22\textwidth]{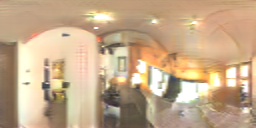} &
		\includegraphics[align=c,width=0.22\textwidth]{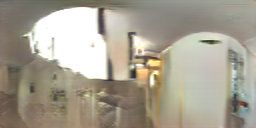} \\
		
		(g) Rotated Ground Truth &
		\includegraphics[align=c,width=0.22\textwidth]{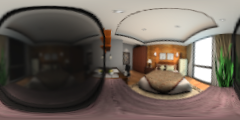} &
		\includegraphics[align=c,width=0.22\textwidth]{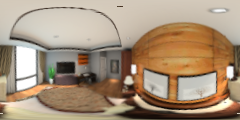} &
		\includegraphics[align=c,width=0.22\textwidth]{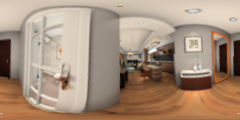} \\
		(h) Corrected Input &
		\includegraphics[align=c,width=0.22\textwidth]{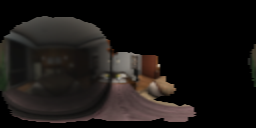} &
		\includegraphics[align=c,width=0.22\textwidth]{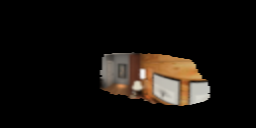} &
		\includegraphics[align=c,width=0.22\textwidth]{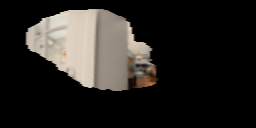} \\
		
		(i) EnvMapNet (Ours on (h)) &
		\includegraphics[align=c,width=0.22\textwidth]{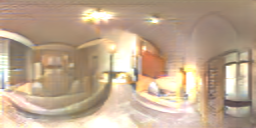} &
		\includegraphics[align=c,width=0.22\textwidth]{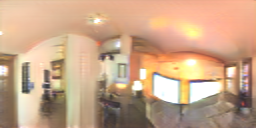} &
		\includegraphics[align=c,width=0.22\textwidth]{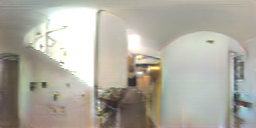} \\
	\end{tabular}
	\caption{Qualitative comparisons with the results from Srinivasan~\etal~\cite{srinivasan20lighthouse} and their re-implementation of \cite{LeGendre:2019:DLI:3306307.3328173} and \cite{song2019neural}. See Appendix~\ref{app_compare}. }\label{app_lighthouse}
\end{figure*}

\section{Comparisons with recent methods} \label{app_compare}
In this section we expand on the qualitative results from paper Section~\ref{qualitative_results} and provide  comparisons to the very recent work of Srinivasan~\etal~\cite{srinivasan20lighthouse}, that trained and evaluated on synthetic LDR images from InteriorNet \cite{InteriorNet18}. The authors also provided results from their re-implementation of Deep Light~\cite{LeGendre:2019:DLI:3306307.3328173} and Neural Illumination~\cite{song2019neural}. We used the images  and results provided as part of their paper and show the comparison to our results in Figure~\ref{app_lighthouse}. We note that the input to our algorithm was only the incomplete panorama shown in Figure~\ref{app_lighthouse}(a), while stereo images were used for Srinivasan~\etal~\cite{srinivasan20lighthouse} (c), and the $32\times32$ sphere image output from LeGendre~\etal~\cite{LeGendre:2019:DLI:3306307.3328173} was converted to an equirectangular projection by the authors (d). Furthermore, compared methods were (re)trained by Srinivasan~\etal on the same synthetic dataset (InteriorNet~\cite{InteriorNet18}), while our results are using the same network as before. It was trained on $2,810$ images from public datasets of real-world images - Laval Indoor HDR dataset~\cite{gardner-sigasia-17}  and PanoContext LDR panoramas~\cite{panocontext} - as  discussed in paper Section~\ref{sec_data}.

Since the network used by Srinivasan~\etal was trained on stereo input and LDR ground truth (with reverse gamma), we only make qualitative comparisons for LDR completion/generation of environment maps. We noticed that the predictions made by the algorithm from Srinivasan~\etal  resembles the (unseen) ground truth textures very closely. For example, the window, curtains and wall boundaries in the first two images shown in Figure~\ref{app_lighthouse}(c) closely match the corresponding regions in Figure~\ref{app_lighthouse}(b). This is surprising given that the input panorama does not contain this information.

Note that our model was trained with images which are aligned with vertical axis being gravity, that is, the horizon is aligned to the horizontal image axis. This is easily achieved in mobile AR using device pose and orientation, and it also avoids ambiguity when training a network. The images provided as input (a) have different rotations resulting in some of the artifacts  observed. Additionally we observe that the input panoramas are not just masked  subset of pixels from the ground truth. The artifacts in the input, such as the blocks and aliasing seen on the left side of the first input, could potentially be from reprojection done by Srinivasan~\etal, which is carried forward by our results as part of `known input'. We manually corrected the above by `straightening' the panoramas (g), and using the input mask to create Corrected Input (h), and show  results from our method using  this corrected input in row (i). Qualitatively we can observe that our method can produce plausible and perceptually pleasing reflections in these new scenes, even though our model was not trained on the same dataset.

\section{Results  and comparison }
\label{sec_appendix_res}
In this section we provide several images to demonstrate the effectiveness of our method, for rendering virtual objects in mobile AR applications, and expand on the images and results discussed in paper ~Section~\ref{sec_quantitative}.

\begin{figure*}
	\centering
	\begin{tabular}{cccc}
		\textbf{Input Crop} &	\textbf{EnvMapNet (Ours)} & \textbf{Ground Truth} & \textbf{Gardner \etal\cite{gardner-sigasia-17}} \\
		
		\includegraphics[align = b, width = 0.1\linewidth]{images/crops/crop_114.png} &
		\includegraphics[align = b, width = 0.1\linewidth]{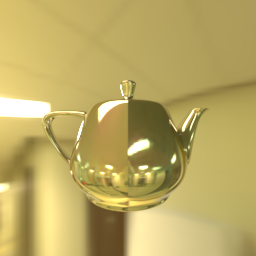}
		\includegraphics[align = b, width = 0.1\linewidth]{images/j9xrpkftr3/rendering_v2/im_dragon_114.png} &
		\includegraphics[align = b, width = 0.1\linewidth]{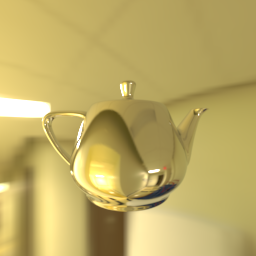}
		\includegraphics[align = b, width = 0.1\linewidth]{images/j9xrpkftr3/rendering_v2/gt_dragon_114.png} &
		\includegraphics[align = b, width = 0.1\linewidth]{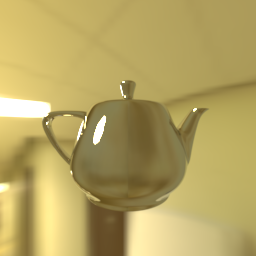}
		\includegraphics[align = b, width = 0.1\linewidth]{images/j9xrpkftr3/rendering_v2/lv_dragon_114.png} \\
		& \shortstack{\includegraphics[align = b, width = 0.2\linewidth]{images/j9xrpkftr3/ldr_114.jpg}\\24.65  } &
		\shortstack{\includegraphics[align = b, width = 0.2\linewidth]{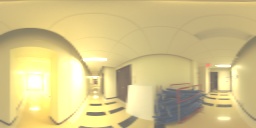}\\ AngularError  } &
		\shortstack{\includegraphics[align = b, width = 0.2\linewidth]{images/laval_results/fullset/output_114.jpg}\\79.96  } \\
		\includegraphics[align = b, width = 0.1\linewidth]{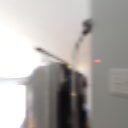} &
		\includegraphics[align = b, width = 0.1\linewidth]{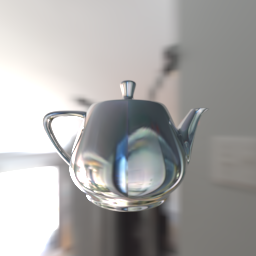}
		\includegraphics[align = b, width = 0.1\linewidth]{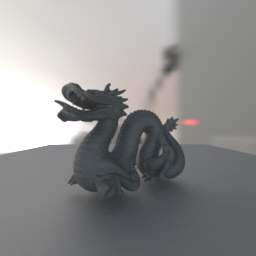} &
		\includegraphics[align = b, width = 0.1\linewidth]{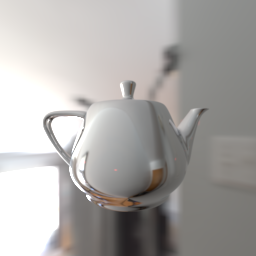}
		\includegraphics[align = b, width = 0.1\linewidth]{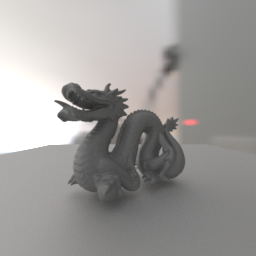} &
		\includegraphics[align = b, width = 0.1\linewidth]{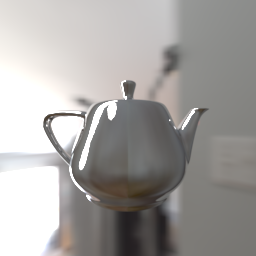}
		\includegraphics[align = b, width = 0.1\linewidth]{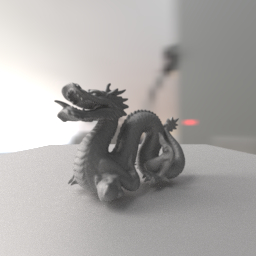} \\
		& \shortstack{\includegraphics[align = b, width = 0.2\linewidth]{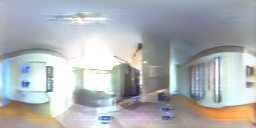}\\8.3  } &
		\shortstack{\includegraphics[align = b, width = 0.2\linewidth]{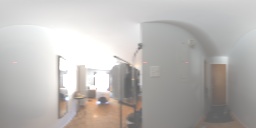}\\ AngularError  } &
		\shortstack{\includegraphics[align = b, width = 0.2\linewidth]{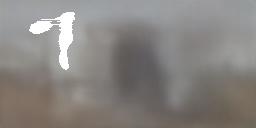}\\68.04  } \\
		\includegraphics[align = b, width = 0.1\linewidth]{images/crops/crop_103.png} &
		\includegraphics[align = b, width = 0.1\linewidth]{images/j9xrpkftr3/rendering_v2/im_teapot2_103.png}
		\includegraphics[align = b, width = 0.1\linewidth]{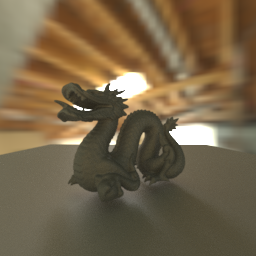} &
		\includegraphics[align = b, width = 0.1\linewidth]{images/j9xrpkftr3/rendering_v2/gt_teapot2_103.png}
		\includegraphics[align = b, width = 0.1\linewidth]{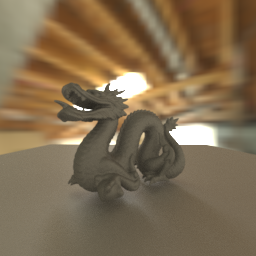} &
		\includegraphics[align = b, width = 0.1\linewidth]{images/j9xrpkftr3/rendering_v2/lv_teapot2_103.png}
		\includegraphics[align = b, width = 0.1\linewidth]{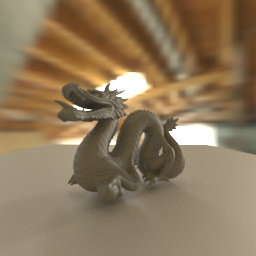} \\
		& \shortstack{\includegraphics[align = b, width = 0.2\linewidth]{images/j9xrpkftr3/ldr_103.jpg}\\1.04  } &
		\shortstack{\includegraphics[align = b, width = 0.2\linewidth]{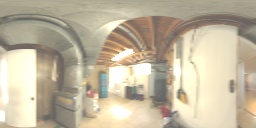}\\ AngularError  } &
		\shortstack{\includegraphics[align = b, width = 0.2\linewidth]{images/laval_results/fullset/output_103.jpg}\\57.93  } \\
		\includegraphics[align = b, width = 0.1\linewidth]{images/crops/crop_97.png} &
		\includegraphics[align = b, width = 0.1\linewidth]{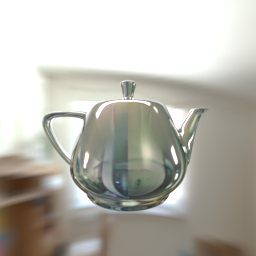}
		\includegraphics[align = b, width = 0.1\linewidth]{images/j9xrpkftr3/rendering_v2/im_dragon_97.png} &
		\includegraphics[align = b, width = 0.1\linewidth]{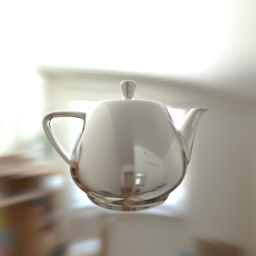}
		\includegraphics[align = b, width = 0.1\linewidth]{images/j9xrpkftr3/rendering_v2/gt_dragon_97.png} &
		\includegraphics[align = b, width = 0.1\linewidth]{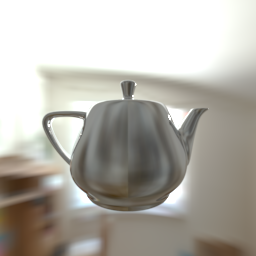}
		\includegraphics[align = b, width = 0.1\linewidth]{images/j9xrpkftr3/rendering_v2/lv_dragon_97.png} \\
		& \shortstack{\includegraphics[align = b, width = 0.2\linewidth]{images/j9xrpkftr3/ldr_97.jpg}\\52.41  } &
		\shortstack{\includegraphics[align = b, width = 0.2\linewidth]{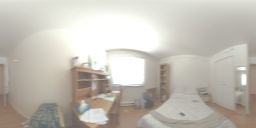}\\ AngularError  } &
		\shortstack{\includegraphics[align = b, width = 0.2\linewidth]{images/laval_results/fullset/output_97.jpg}\\50.99  } \\

		\includegraphics[align = b, width = 0.1\linewidth]{images/crops/crop_243.png} &
		\includegraphics[align = b, width = 0.1\linewidth]{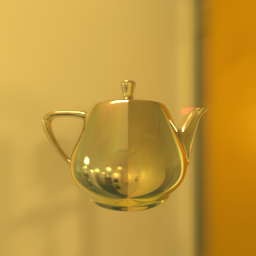}
		\includegraphics[align = b, width = 0.1\linewidth]{images/j9xrpkftr3/rendering_v2/im_dragon_243.png} &
		\includegraphics[align = b, width = 0.1\linewidth]{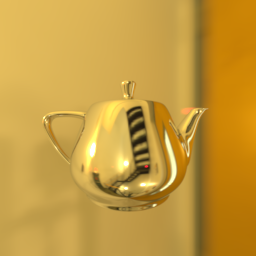}
		\includegraphics[align = b, width = 0.1\linewidth]{images/j9xrpkftr3/rendering_v2/gt_dragon_243.png} &
		\includegraphics[align = b, width = 0.1\linewidth]{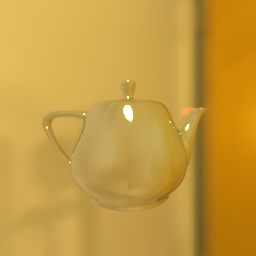}
		\includegraphics[align = b, width = 0.1\linewidth]{images/j9xrpkftr3/rendering_v2/lv_dragon_243.png} \\
		& \shortstack{\includegraphics[align = b, width = 0.2\linewidth]{images/j9xrpkftr3/ldr_243.jpg}\\35.33  } &
		\shortstack{\includegraphics[align = b, width = 0.2\linewidth]{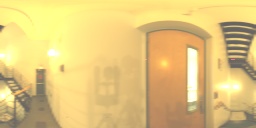}\\ AngularError  } &
		\shortstack{\includegraphics[align = b, width = 0.2\linewidth]{images/laval_results/fullset/output_243.jpg}\\57.67  } \\

	\end{tabular}
	\caption{Results with variety of angular errors. Each result is shown over a  row containing the input crop, rendered images with reflective teapot and diffuse dragon, with corresponding environment maps below each. As discussed in  paper Section~\ref{sec_results}  our method is qualitatively and quantitatively able to produce visually coherent environment maps for  lighting, shadows  and reflection.}
	\label{fig_selected_res}
\end{figure*}

\begin{figure*}
	\centering
	\begin{tabular}{cccc}
		\textbf{Input Crop} & \textbf{EnvMapNet (Ours)} & \textbf{Ground Truth} & \textbf{Gardner \etal\cite{gardner-sigasia-17}} \\
		\includegraphics[align = b, width = 0.1\linewidth]{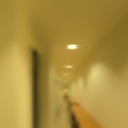} &
		\includegraphics[align = b, width = 0.1\linewidth]{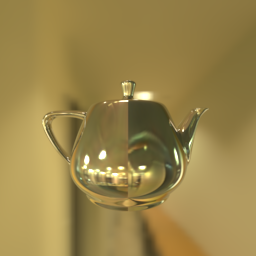}
		\includegraphics[align = b, width = 0.1\linewidth]{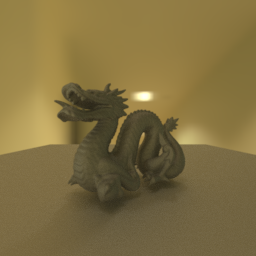} &
		\includegraphics[align = b, width = 0.1\linewidth]{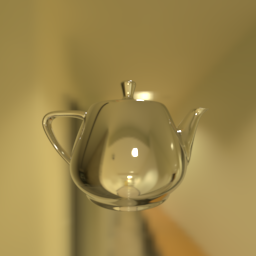}
		\includegraphics[align = b, width = 0.1\linewidth]{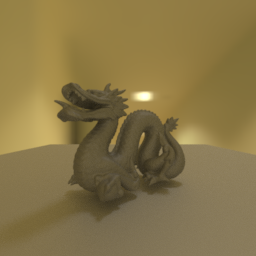} &
		\includegraphics[align = b, width = 0.1\linewidth]{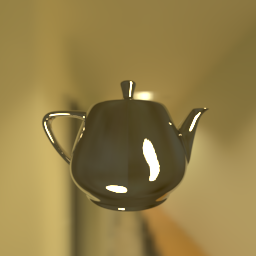}
		\includegraphics[align = b, width = 0.1\linewidth]{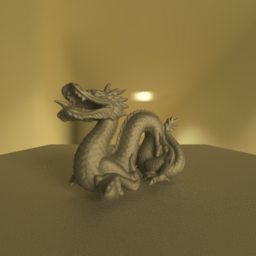} \\
		& \shortstack{\includegraphics[align = b, width = 0.2\linewidth]{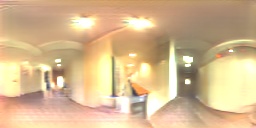}\\37.53  } &
		\shortstack{\includegraphics[align = b, width = 0.2\linewidth]{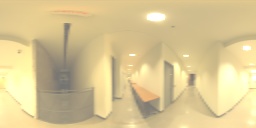}\\ AngularError  } &
		\shortstack{\includegraphics[align = b, width = 0.2\linewidth]{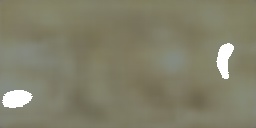}\\48.42  } \\
		\includegraphics[align = b, width = 0.1\linewidth]{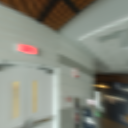} &
		\includegraphics[align = b, width = 0.1\linewidth]{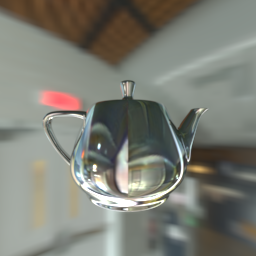}
		\includegraphics[align = b, width = 0.1\linewidth]{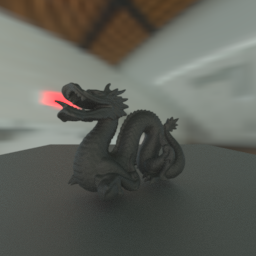} &
		\includegraphics[align = b, width = 0.1\linewidth]{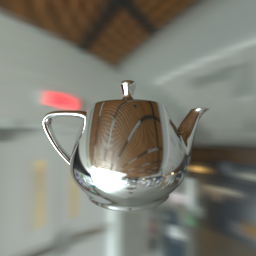}
		\includegraphics[align = b, width = 0.1\linewidth]{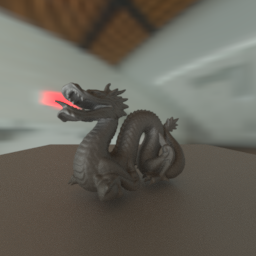} &
		\includegraphics[align = b, width = 0.1\linewidth]{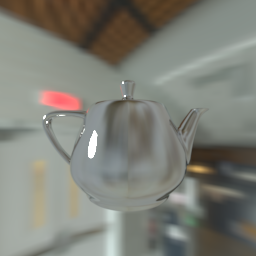}
		\includegraphics[align = b, width = 0.1\linewidth]{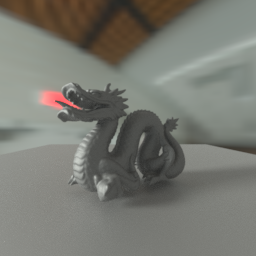} \\
		& \shortstack{\includegraphics[align = b, width = 0.2\linewidth]{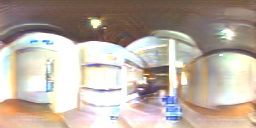}\\40.18  } &
		\shortstack{\includegraphics[align = b, width = 0.2\linewidth]{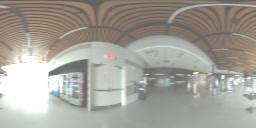}\\ AngularError  } &
		\shortstack{\includegraphics[align = b, width = 0.2\linewidth]{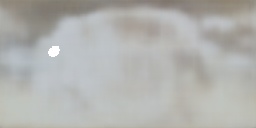}\\53.26  }\\
		\includegraphics[align = b, width = 0.1\linewidth]{images/crops/crop_111.png} &
		\includegraphics[align = b, width = 0.1\linewidth]{images/j9xrpkftr3/rendering_v2/im_teapot2_111.png}
		\includegraphics[align = b, width = 0.1\linewidth]{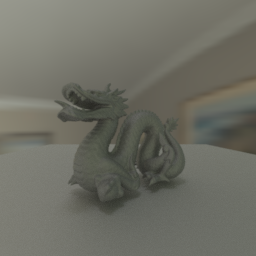} &
		\includegraphics[align = b, width = 0.1\linewidth]{images/j9xrpkftr3/rendering_v2/gt_teapot2_111.png}
		\includegraphics[align = b, width = 0.1\linewidth]{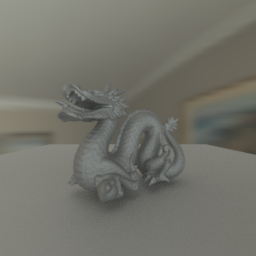} &
		\includegraphics[align = b, width = 0.1\linewidth]{images/j9xrpkftr3/rendering_v2/lv_teapot2_111.png}
		\includegraphics[align = b, width = 0.1\linewidth]{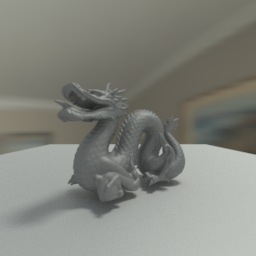} \\
		& \shortstack{\includegraphics[align = b, width = 0.2\linewidth]{images/j9xrpkftr3/ldr_111.jpg}\\41.54  } &
		\shortstack{\includegraphics[align = b, width = 0.2\linewidth]{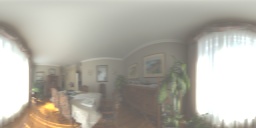}\\ AngularError  } &
		\shortstack{\includegraphics[align = b, width = 0.2\linewidth]{images/laval_results/fullset/output_111.jpg}\\111.82  } \\
		\includegraphics[align = b, width = 0.1\linewidth]{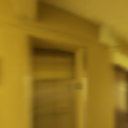} &
		\includegraphics[align = b, width = 0.1\linewidth]{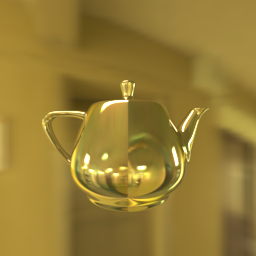}
		\includegraphics[align = b, width = 0.1\linewidth]{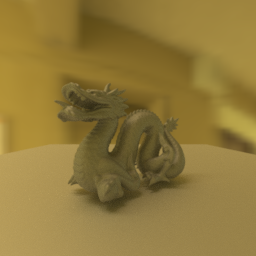} &
		\includegraphics[align = b, width = 0.1\linewidth]{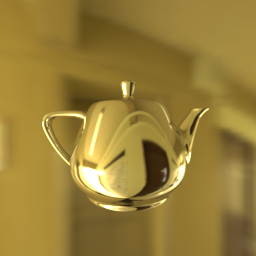}
		\includegraphics[align = b, width = 0.1\linewidth]{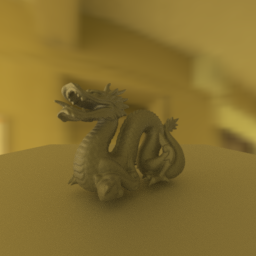} &
		\includegraphics[align = b, width = 0.1\linewidth]{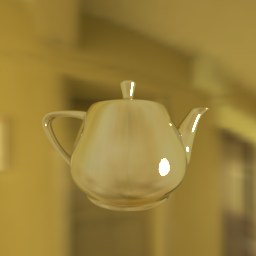}
		\includegraphics[align = b, width = 0.1\linewidth]{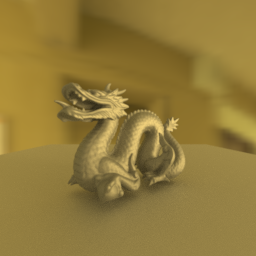} \\
		& \shortstack{\includegraphics[align = b, width = 0.2\linewidth]{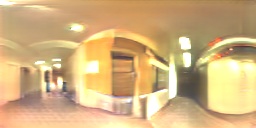}\\41.59  } &
		\shortstack{\includegraphics[align = b, width = 0.2\linewidth]{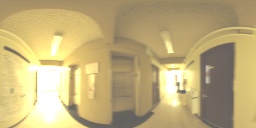}\\ AngularError  } &
		\shortstack{\includegraphics[align = b, width = 0.2\linewidth]{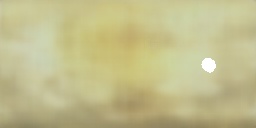}\\90.54  }\\
		\includegraphics[align = b, width = 0.1\linewidth]{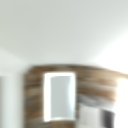} &
		\includegraphics[align = b, width = 0.1\linewidth]{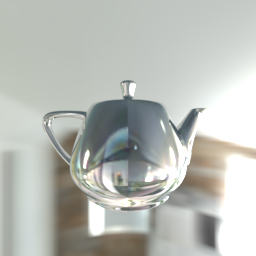}
		\includegraphics[align = b, width = 0.1\linewidth]{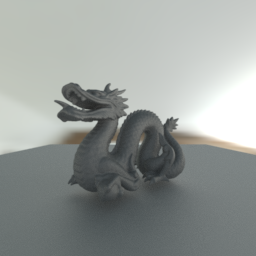} &
		\includegraphics[align = b, width = 0.1\linewidth]{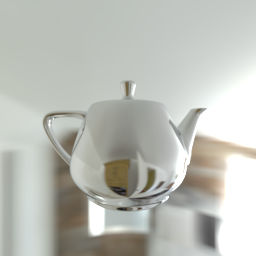}
		\includegraphics[align = b, width = 0.1\linewidth]{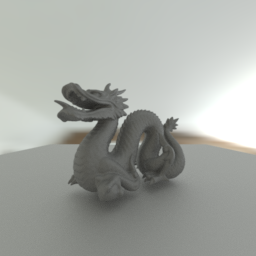} &
		\includegraphics[align = b, width = 0.1\linewidth]{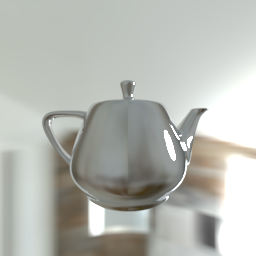}
		\includegraphics[align = b, width = 0.1\linewidth]{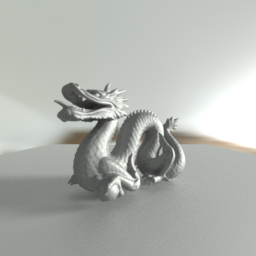} \\
		& \shortstack{\includegraphics[align = b, width = 0.2\linewidth]{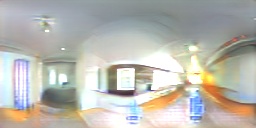}\\34.5  } &
		\shortstack{\includegraphics[align = b, width = 0.2\linewidth]{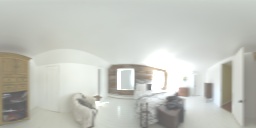}\\ AngularError  } &
		\shortstack{\includegraphics[align = b, width = 0.2\linewidth]{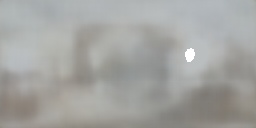}\\56.83  } \\
	\end{tabular}
	\caption{More results with variety of angular errors. Each result is shown over a row containing the input crop, rendered images with reflective teapot and diffuse dragon, with corresponding environment maps below each. See Figure~\ref{fig_selected_res} and paper Section~\ref{sec_results}.}\label{fig_selected_res2}
\end{figure*}

 In Figures~\ref{fig_selected_res} and ~\ref{fig_selected_res2} we show predicted environment maps from Gardner~\etal~\cite{gardner-sigasia-17} and our method over a variety of scenes and lighting conditions. We further show their use for rendering both reflective (teapot) and diffuse (dragon) objects\footnote{http://graphics.stanford.edu/data/3Dscanrep/} with shadows. Each result is shown over a pair of rows containing the input crop, rendered images, and  corresponding environment maps with angular error below each. To enable fair and future comparison, we crop the test environment maps in the center for $90$ degree FoV. We rectify and provide the \textbf{Input Crop} to  \cite{gardner-sigasia-17}, and project them to a equirectangular map for our method. We obtain the predicted equirectangular maps from each image, extract the parametric lights for calculation of AngularError as described in paper Section~\ref{sec_metrics}, and proceed to use for rendering. 

We use Tungsten\footnote{https://github.com/tunabrain/tungsten} an open source path tracer, to generate renderings of an aluminum metal finish teapot and a lambertian material on the dragon objects. Only for the purpose of rendering, we perform the following post-processing operations on each predicted equirectangular map. Since each method provides the result in an arbitrary intensity range, we scale the pixel intensities for each predicted result such that the average intensity in the region provided as input is equal to that of the ground truth HDR. We further overlay the pixels from the ground truth corresponding to the input FoV, to simulate the AR video backdrop and common background for each rendering. We provide this equirectangular map as the input for Tungsten, position the virtual camera to match the input crop provided, and obtain the rendered images shown in Figures~\ref{fig_selected_res} and ~\ref{fig_selected_res2} for \textbf{EnvMapNet (Ours), Ground Truth} and \textbf{Gardner~\etal~\cite{gardner-sigasia-17}}. We stress that this post-processing was only done for the offline rendering pipeline and not for other comparisons.

As previously detailed, this simulates that a virtual object is placed at the center of a probe which is illuminated using the environment map.  As shown in the images, our results generate perceptually pleasing and accurate lighting for the virtual object compared to those from \cite{gardner-sigasia-17}, for both material finishes.  Though our method cannot (and should not) create the unseen ground truth scene exactly, the level of detail generated is clearly plausible and provides a better visual coherence compared to those from ~\cite{gardner-sigasia-17}. This aspect regarding image resolution and quality of the details is captured by the FID metric as detailed in the main paper. 
The renderings of the diffuse dragon object on a plane highlight the difference in accuracy of estimated light direction, as captured by our AngularError metric listed below each result. The images show estimated environment maps with a wide range of  errors from both methods, and as can be observed, overall our estimates produce shadows which better match those from the ground truth environment map.

{\small
	\bibliographystyle{ieee_fullname}
	\bibliography{EnvMapNet_bib}
}

\makeatletter\@input{EnvMapNet_cvpraux.tex}\makeatother